\documentclass{article}

\PassOptionsToPackage{numbers, compress}{natbib}

\usepackage[preprint]{neurips_2024}

\usepackage{booktabs}
\usepackage{graphicx}
\usepackage{capt-of}
\usepackage[utf8]{inputenc} 
\usepackage[T1]{fontenc}    
\usepackage{hyperref}       
\usepackage{url}            
\usepackage{booktabs}       
\usepackage{amsfonts}       
\usepackage{nicefrac}       
\usepackage{microtype}      
\usepackage{xcolor}         
\usepackage{multirow}
\usepackage{amsmath}
\usepackage{tabularx}
\usepackage{pifont}
\usepackage{bbding}  
\usepackage{wrapfig}
\RequirePackage{subcaption}

\title{Towards Ultra-High-Definition Image Deraining: A Benchmark and An Efficient Method}

\author{%
    Hongming Chen$^1$\thanks{Authors contributed equally to this work.}, Xiang Chen$^2$\footnotemark[1], Chen Wu$^{3}$, Zhuoran Zheng$^{4}$, Jinshan Pan$^2$, Xianping Fu$^1$ \\
    $^1$Dalian Martime University,
    $^2$Nanjing University of Science and Technology \\
    $^3$University of Science and Technology of China,
    $^4$Sun Yat-sen University\
}

\begin{document}

\maketitle

\begin{abstract}
Despite significant progress has been made in image deraining, existing approaches are mostly carried out on low-resolution images. The effectiveness of these methods on high-resolution images is still unknown, especially for ultra-high-definition (UHD) images, given the continuous advancement of imaging devices. 
In this paper, we focus on the task of UHD image deraining, and contribute the first large-scale UHD image deraining dataset, 4K-Rain13k, that contains 13,000 image pairs at 4K resolution. Based on this dataset, we conduct a benchmark study on existing methods for processing UHD images.
Furthermore, we develop an effective and efficient vision MLP-based architecture (UDR-Mixer) to better solve this task. Specifically, our method contains two building components: a spatial feature rearrangement layer that captures long-range information of UHD images, and a frequency feature modulation layer that facilitates high-quality UHD image reconstruction.
Extensive experimental results demonstrate that our method performs favorably against the state-of-the-art approaches while maintaining a lower model complexity.
The code and dataset will be available at \url{https://github.com/cschenxiang/UDR-Mixer}.
\end{abstract}

\section{Introduction}
\label{sec:intro}
Single image deraining aims to remove the undesired degradation induced by rain streaks from input images, enhancing its visual quality and improving the accuracy of perception system~\cite{chen2022unpaired}.
In image deraininig, deep learning-based methods become predominate ones as the formation of image deraining is quite simplified compared to the conventional prior-based methods~\cite{luo2015removing}. One can choose deep models based on convolutional neural network (CNN)~\cite{jiang2020multi,zamir2021multi,yi2021structure} or Transformer architectures~\cite{xiao2022image,chen2023learning,chen2023sparse,chen2024bidirectional} to directly estimate clear image from rainy one. 

\begin{figure*}[!t]
\centering 	
\begin{subfigure}[t]{0.46\textwidth}
\centering
\includegraphics[width=\textwidth]{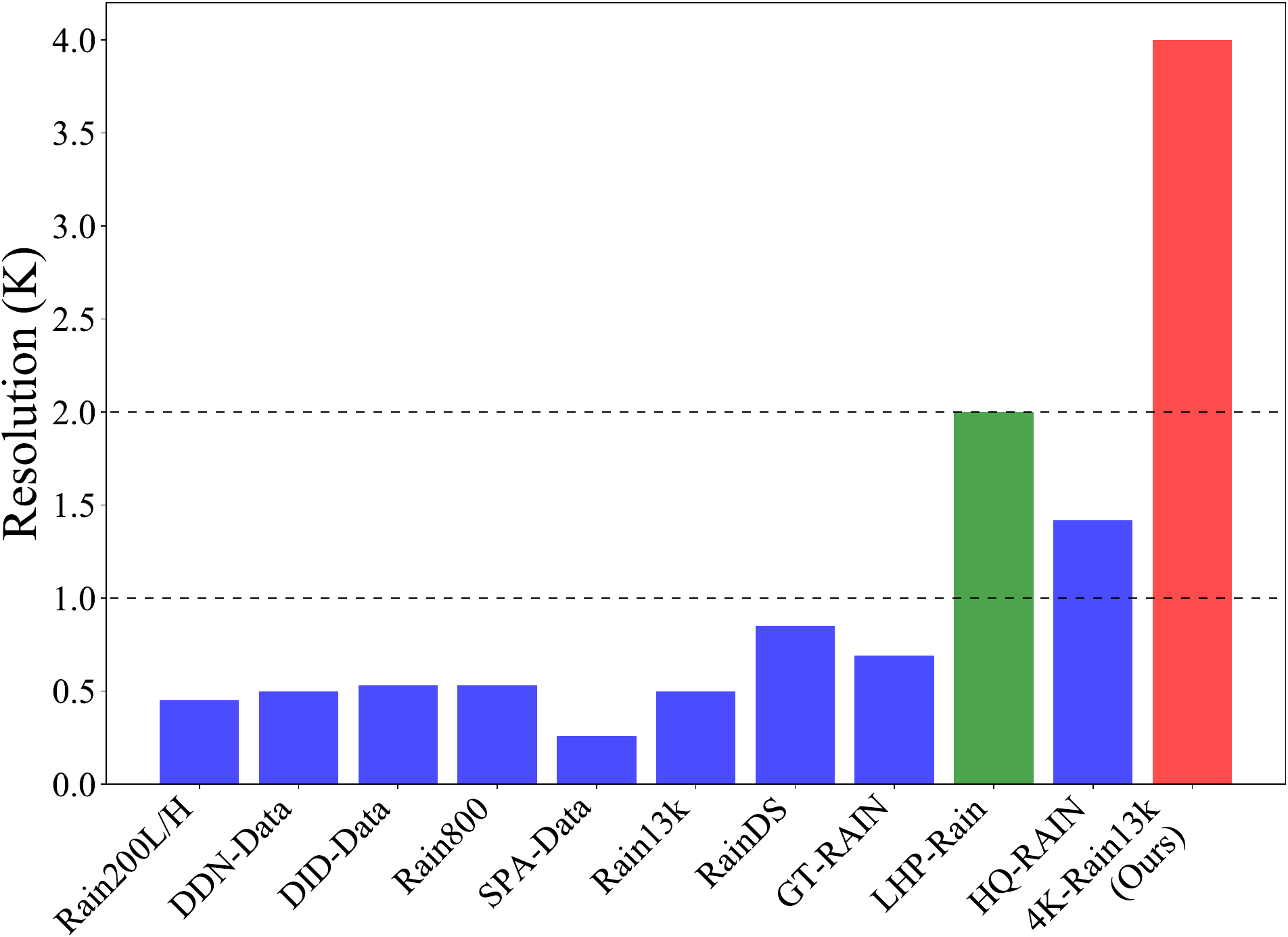}
\caption{Statistics on the image resolution}
\end{subfigure}
\begin{subfigure}[t]{0.53\textwidth}
\centering
\includegraphics[width=\textwidth]{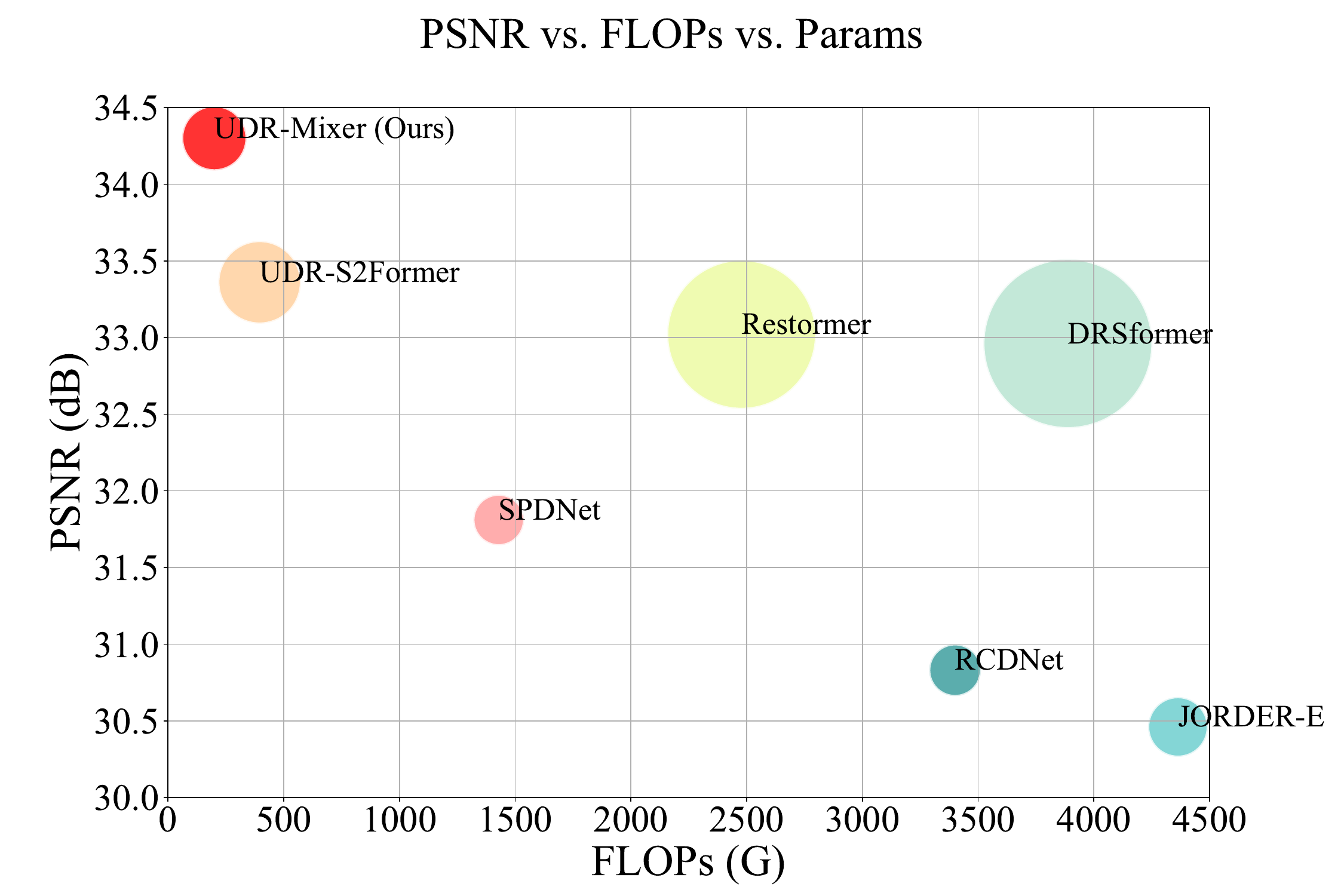}
\caption{Analysis of model computational complexity}
\end{subfigure}
\caption{Comparisons on different benchmarks and methods. (a) As existing datasets have not explored high-resolution images, particularly UHD images, our proposed 4K-Rain13k dataset will fill the gap in this research. (b) Model complexity and performance comparisons of the proposed method and other state-of-the-art models on the proposed 4K-Rain13k dataset in terms of PSNR, model parameters and FLOPs. The area of each circle denotes the number of model parameters. Since most approaches are unable to directly process UHD images, FLOP calculation is based on image sizes of $1024 \times 1024$. Our method achieves a better trade-off between efficiency and performance.}
\label{fig1}
\vspace{-5mm}
\end{figure*}

Among these approaches, most of them are trained and evaluated on low-resolution datasets~\cite{chen2023towards}. These commonly used benchmark datasets consist of 1K or even lower resolution images, such as Rain200L/H~\cite{yang2017deep} and Rain13k~\cite{jiang2020multi}, as illustrated in Figure \ref{fig1}(a). 
Based on the existing empirical studies~\cite{zhang2021benchmarking,wang2023ultra,li2023embedding}, existing image deraining approaches trained on these low-resolution datasets are not likely to generalize well on high-resolution images.
However, few effort has been made in ultra-high definition (UHD) image deraining due to the absence of UHD deraining dataset.
As UHD devices have been widely used, it is urgent and essential to build a high-resolution benchmark and pave the way for future research in this field.

To explore the performance of existing approaches on UHD images, in this paper, we first establish a large-scale dataset called 4K-Rain13k to benchmark existing methods. The proposed 4K-Rain13k contains 13,000 rainy/rain-free image pairs at 4K resolution ($3840 \times 2160$), with 12,500 pairs allocated for training and 500 pairs for testing.
Unlike existing datasets~\cite{yang2017deep,fu2017removing,zhang2018density,zhang2019image,jiang2020multi,liu2021unpaired} that directly add rain streaks proportionally to clear images to synthesize rainy images, we observe that geometric inconsistencies in the lengths and thicknesses of rain streaks between low-resolution and high-resolution rainy images.
In high-resolution UHD images, rain streaks typically appear longer and slighter due to the increased pixel information, whereas in low-resolution images, the same length of rain streaks may appear blurred into shorter or thicker lines due to fewer pixels. To this end, by integrating geometric transformations (i.e., scaling) into the rain synthesis pipeline, we can enhance the harmony and consistency of the synthesized images, enabling them to better reflect the attributes of the original UHD image content. Based on this new dataset, we conduct extensive evaluations to analyze the performance of existing methods.

Furthermore, we find that when dealing with UHD images, most state-of-the-art (SOTA) methods often encounter memory overflow issues, making it difficult to perform full-resolution inference on consumer-grade GPUs.
This motivates us to develop an effective and efficient method tailored for UHD image deraining. In this paper, we develop an vision MLP-based architecture (called UDR-Mixer) to better solve this task, rather than relying on the self-attention mechanism that is computationally expensive in Transformers~\cite{zamir2022restormer}.

The proposed UDR-Mixer consists of two parallel branches, each dedicated to exploring spatial and frequency representations to complement each other. In the main branch, we construct spatial feature mixing blocks (SFMB) as the core components, establishing global information perception through a simple yet effective feature rearrangement mechanism. Unlike strategies based on single-view spatial region rearrangement~\cite{guo2022hire,yu2022s2}, we recursively encode the entire image from different perspectives in multi-stage dimensional transformations and correlate multi-view features by permuting the tensor to better capture long-range pixel dependencies in UHD images. Simultaneously, we introduce an auxiliary branch composed of frequency feature mixing blocks (FFMB) to facilitate high-quality restoration of UHD images. Figure \ref{fig1}(b) illustrates that our method achieves favorable performance with a better trade-off between efficiency and performance. 

The main contributions of this paper are summarized as follows:
\begin{itemize}
\item We propose the first high-quality UHD image deraining dataset (4K-Rain13k). Based on this dataset, we conduct a benchmark evaluation on existing methods for processing UHD images.

\item We develop the spatial feature mixing block and the frequency feature mixing block to handle UHD images efficiently and formulate them into an end-to-end trainable MLP-based network (UDR-Mixer) based on a dual-branch architecture for UHD image deraining.

\item We quantitatively and qualitatively evaluate the proposed method on the proposed 4K-Rain13k dataset as well as real-world UHD images. The results demonstrate that our approach achieves a favorable trade-off between performance and model complexity. 
\end{itemize}

\section{Related Work}
\label{sec:work}
\vspace{-2mm}
{\flushleft\textbf{Single image deraining}.}
When we revisit this field of single image deraining, numerous deraining approaches and benchmark datasets have been proposed in recent years with demonstrated success~\cite{chen2023towards}. Several classic benchmark datasets are widely adopted to evaluate single image deraining performance, such as Rain200L/H~\cite{yang2017deep}, DID-Data~\cite{zhang2018density}, DDN-Data~\cite{fu2017removing} and Rain13k~\cite{jiang2020multi}. These early benchmark datasets consist of lower resolution images (1K or less). However, in this field, there is a lack of exploration specifically for higher resolution images, particularly UHD images. Furthermore, when dealing with UHD images, existing SOTA methods frequently encounter memory overflow issues, preventing them from conducting full-resolution inference on consumer-grade GPUs.

\vspace{-1mm}

{\flushleft\textbf{UHD image processing}.}
With the development of photography equipment, UHD image processing has emerged as a new research trend in recent years~\cite{zheng2021ultra,yu2022towards,wu2024mixnet}. Zheng \emph{et al.}~\cite{zheng2021ultra} formulated the UHD image dehazing network using multi-guide bilateral learning. Zhang \emph{et al.} \cite{zhang2021benchmarking} explored the task of image super-resolution for UHD resolutions, and further created two large-scale image datasets, UHDSR4K and UHDSR8K. Ren \emph{et al.}~\cite{ren2023fast} developed a multi-scale separable network to address UHD deblurring problem. The UHD low-light image enhancement task has also received increasing attention from researchers, and representative datasets include UHD-LOL~\cite{wang2023ultra} and UHD-LL~\cite{li2023embedding}. Beyond that, other related tasks have focused on the application of UHD images, e.g., reflection removal and HDR reconstruction~\cite{zheng2021ultra}. To the best of our knowledge, we first focus on the task of removing rain from UHD images, and we propose both a benchmark dataset and a baseline method.

\vspace{-1mm}

{\flushleft\textbf{Vision MLP}.}
Given the high computational cost of self-attention mechanism in vision Transformers (ViT), several researchers have designed efficient vision models comprising solely multi-layer perceptrons (MLPs). For example, MLP-Mixer~\cite{tolstikhin2021mlp} utilizes a straightforward token-mixing MLP instead of self-attention in ViT, leading to an all-MLP network. It employs token-mixing MLP to capture token relationships and channel-mixing MLP to capture channel relationships. Afterwards, some studies further improve the performance of MLP-based models by designing other architectures, such as gMLP~\cite{liu2021pay} and Hire-MLP~\cite{guo2022hire}. Recently, Tu \emph{et al.}~\cite{tu2022maxim} formulated a multi-axis MLP-based framework MAXIM for image processing tasks. Wu \emph{et al.}~\cite{wu2024mixnet} developed an efficient MixNet for UHD low-light image enhancement by modeling global and local feature dependencies. Inspired by these works, we leverage the vision MLP architecture to flexibly handle UHD image deraining.

\vspace{-2mm}

\section{UHD Image Deraining Dataset Construction}
\vspace{-2mm}
To evaluate the performance of existing approaches on the UHD image deraining problem, we first create a large-scale benchmark dataset named 4K-Rain13k. 
We note that existing low-resolution rain datasets~\cite{yang2017deep,fu2017removing,zhang2018density,zhang2019image,jiang2020multi,liu2021unpaired} simply add rain streaks into the clear backgrounds to obtain rainy images. 
However, this copy-and-pasting approach is not suitable for synthesizing UHD rainy images due to the \emph{geometric inconsistency} between the low-resolution and high-resolution image synthesis processes. 
Thus, we develop an effective method for synthesizing rainy images tailored for UHD images, aiming to achieve more realistic visual effects. Our method involves background collection, rain streak generation and geometric transformation, which will be presented below.

\vspace{-1mm}

{\flushleft\textbf{Background collection}.}
We collect numerous clear backgrounds using a Python program based on Scrapy to download high-resolution images from the web and various devices. Our ground-truths includes a wide range of typical daytime and nighttime scenes in urban locations (e.g., buildings, streets, cityscapes) as well as natural landscapes (e.g., lakes, hills, and vegetations).

\vspace{-1mm}

{\flushleft\textbf{Rain streak generation}.}
The diversity and fidelity of rain play crucial roles in the synthesis of rain streaks. Instead of using Photoshop software to render rain streaks, we synthesize corresponding rainy images by modeling the generation of rain streaks as a motion blur process to ensure diversity. In addition, we take into account the transparency of rain layer to ensure fidelity through alpha blending.

\vspace{-1mm}

{\flushleft\textbf{Geometric transformation}.}
In fact, there is an easily overlooked problem of geometric inconsistency in the synthesis of low resolution and high-resolution rainy images, with noticeable discrepancies in the length and thickness of rain streaks.
Specifically, in high-resolution 4K images, with more pixel information available, rain streaks tend to appear longer and slighter as each rain-effected region can be accurately represented. In low-resolution images with fewer pixels, rain streaks of the same length may appear shorter or thicker due to blurring, resulting in a rougher and more ambiguous appearance.

\begin{figure*}[t]
	\begin{minipage}{0.59\linewidth}
		\centering
		\captionof{table}{\small{
				Comparison between existing image deraining datasets and our proposed 4K-Rain13k dataset. `Number': the number of paired images. `Resolution': the average resolution of the dataset.}}
		\scalebox{0.9}{
			\begin{tabular}{cccc}
				\hline
				Dataset           & Year & Number & Avg. Resolution \\ \hline
				Rain200L/H~\cite{yang2017deep}        & 2017 & 2.0K     &   $435 \times 366$          \\
				DDN-Data~\cite{fu2017removing}          & 2017 & 13.0K    &   $489 \times 428$    \\
				DID-Data~\cite{zhang2018density}          & 2018 & 13.2K  &    $512 \times 512$        \\
				Rain800~\cite{zhang2019image}           & 2019 & 0.8K   &  $518 \times 419$        \\
				SPA-Data~\cite{wang2019spatial}          & 2019 & 29.5K  & $256 \times 256$         \\
				Rain13k~\cite{jiang2020multi}           & 2020 & 13.7K  &   $482 \times 419$    \\
				RainDirection~\cite{liu2021unpaired}     & 2021 & 3.3K  &     $1945 \times 1444$      \\\				    RainDS~\cite{quan2021removing}           & 2021 & 1.4K  & $818 \times 460$         \\
				GT-RAIN~\cite{ba2022not}           & 2022 & 31.5K  & $666 \times 339$         \\
				LHP-Rain~\cite{guo2023sky}          & 2023 & 1.0M   & $1920 \times 1080$       \\
				HQ-RAIN~\cite{chen2023towards}           & 2023 & 5.0K   & $1367 \times 931$        \\
				4K-Rain13k (Ours) & 2024 & 13.0K    & $3840 \times 2160$       \\ \hline
			\end{tabular}
		}
		\label{table1}
	\end{minipage}
	\hfill
	\begin{minipage}{0.39\linewidth}
		\centering
		\includegraphics[width=1.0\linewidth]{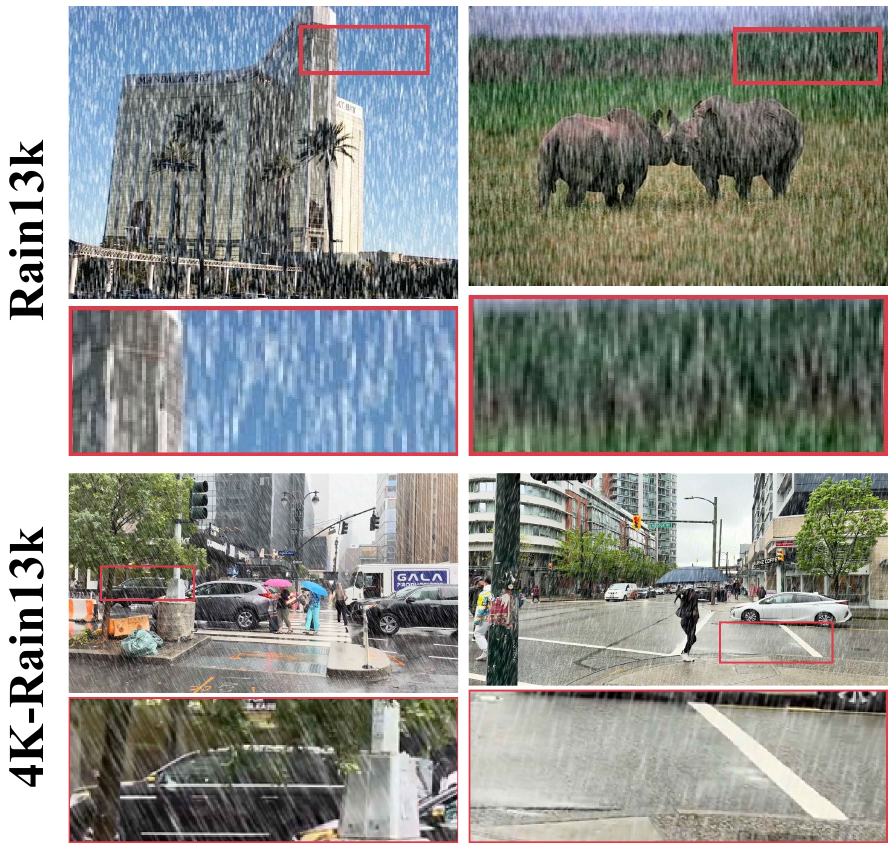}
		\caption{\small{Sample images from the Rain13k dataset~\cite{jiang2020multi} and our 4K-Rain13k dataset.}}
		\label{fig2}
	\end{minipage}
 \vspace{-4mm}
\end{figure*}

To this end, we introduce geometric transformation operations to adjust the scale of rain streaks on the UHD images. By applying simple geometric transformations such as scaling, we aim to harmonize the proportions and sizes of rain streaks in the synthesized images with those observed in the high-resolution 4K images. This step helps alleviate the geometric disparities caused by varying image resolutions, ensuring that the rain streak patterns maintain their intended appearance and spatial relationships during the image synthesis process. We present sample images in Figure~\ref{fig2}.

\vspace{-1mm}

{\flushleft\textbf{Benchmark statistics}.}
Our proposed 4K-Rain13k dataset contains 12,500 synthetic training pairs and 500 test images at 4K resolution ($3840 \times 2160$). The training and test partitions are distinct in terms of their scenes and data, with no overlap. Table \ref{table1} presents a comparison between our dataset and existing public datasets. Following~\cite{chen2023towards}, we utilize the Kullback-Leibler divergence (KLD), also known as relative entropy, to quantify the difference between the distribution of synthetic images and real images. Due to limited space, we provide the analysis results in the \emph{supplementary material}. The results show that our 4K-Rain13k dataset is closer to the distribution of real-world rainy scenes.

\vspace{-2mm}

\section{Proposed UDR-Mixer} 
\vspace{-2mm}
In this section, we develop an effective and efficient MLP-based method (UDR-Mixer) for UHD image deraining. We first describe the overall pipeline, and then present the details of two main components, i.e., spatial feature mixing blocks (SFMB) and frequency feature mixing blocks (FFMB). 

\subsection{Overall pipeline}
As shown in Figure \ref{fig3}, UDR-Mixer is an end-to-end dual-branch parallel network architecture that models UHD images by exploring both spatial and frequency domain information. Specifically, we first embed an input rainy image $I_{rain} \in \mathbb{R}^{H \times W \times 3}$ into the feature space $F_0 \in \mathbb{R}^{H \times W \times C}$ through a $3 \times 3$ convolution layer, where $H$, $W$ and $C$ represents the height, width, and channel, respectively.
To reduce computational complexity in high-resolution images, following previous studies~\cite{li2023embedding,wu2024mixnet}, we employ a PixelUnshuffle operation to downsample the features to $1/4$ of the original resolution. Then, the low-level features are processed by an encoder-decoder network consisting of $N_{i}$ SFMBs with a $2 \times$ downsample operation and a $2 \times$ upsample operation to produce output features.

To alleviate the issue of losing image details caused by  straightforward downsampling operations, we further introduce an auxiliary branch to help UHD image reconstruction. Specifically, we stack $N_{i}$ FFMBs to excavate the frequency information of the full-resolution UHD image. Then, the learned deep features are fed to the decoder network of the main branch for guiding latent clear image restoration. Finally, the output features are obtained to estimate the derained image using a $3 \times 3$ convolutional layer followed by a PixelShuffle operation. We employ the $L_1$ loss as the objective function:
\begin{equation}
\mathcal{L}=\left\|I_{derain}-I_{g t}\right\|_1,
\end{equation}
where $I_{g t}$ denotes the ground-truth image, and $\|\cdot\|_1$ denotes the $L_1$-norm. 

\subsection{Spatial feature mixing block}
Inspired by MLP-Mixer~\cite{tolstikhin2021mlp}, we introduce the MLP-based backbone module to encode feature information. Firstly, we develop a simple yet effective SFMB to aggregate global spatial information. Given the input feature $\mathbf{X}_{l-1} \in \mathbb{R}^{H \times W \times C}$, the proposed SFMB can be formulated as:
\begin{equation}
\begin{aligned}
& \mathbf{X}_l^{\prime}=\mathbf{X}_{l-1}+\text{SFRL}\left(\text{LN}\left(\mathbf{X}_{l-1}\right)\right), \\
& \mathbf{X}_l=\mathbf{X}_l^{\prime}+\text{FFL}\left(\text{LN}\left(\mathbf{X}_l^{\prime}\right)\right), 
\end{aligned}
\end{equation}
where LN refers to the layer normalization. $\mathbf{X}_l^{\prime}$ and $\mathbf{X}_l$ are output feature from the spatial feature rearrangement layer (SFRL) and feed-forward layer (FFL).

\begin{figure}[!t]
\centering
\includegraphics[width=0.95\textwidth]{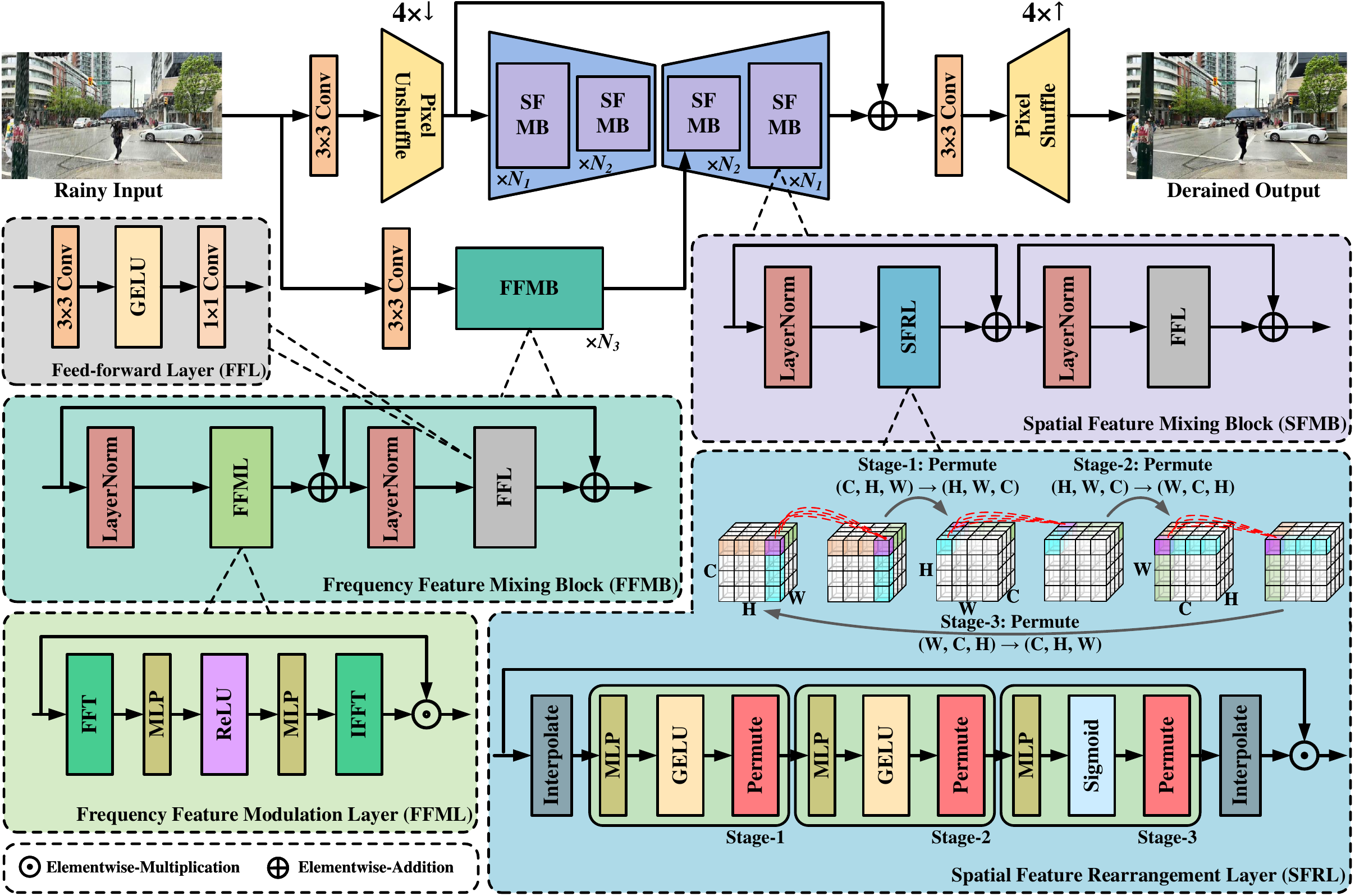}
\caption{The overall architecture of the proposed UDR-Mixer for UHD image deraining, which mainly contains pixel unshuffle/shuffle operations and MLP-based components, i.e., spatial feature mixing blocks (SFMB) and frequency feature mixing blocks (FFMB).}
\label{fig3}
\vspace{-4mm}
\end{figure}

\vspace{-1mm}

{\flushleft\textbf{Spatial feature rearrangement layer}.}
To model global spatial information with lower computational costs, we develop a SFRL based on continuous rearrangements and channel-mixing MLPs. Different from Hire-MLP~\cite{guo2022hire} that utilizes region rearrangement strategy, we introduce a more flexible mechanism via dimension transformation operations, which directly considers the spatial properties between pixels in 3D feature maps. This enables the model to progressively capture global features across the entire image by scrolling, thereby establishing a gradual perception of long-range information from UHD images (see Figure~\ref{fig3}).
Specifically, we first normalize the input features, and then perform multi-stage dimension transformations to rotate the spatial perspective of the tensor across three dimensions of $H$, $W$ and $C$. Here, the 3D feature map undergoes recursive encoding from $(C, H, W)$ to $(H, W, C)$ and then to $(W, C, H)$, enabling the capture of global spatial information through multi-view dimensions.
Finally, we adjust the feature map to the original resolution, and interact with the input features to activate useful features. Mathematically, given an input feature $\mathbf{F}_{0}$, the feature propagation process of SFRL can be expressed as:
\begin{equation}
\begin{aligned}
& \mathbf{F}_l=\text{Interpolate}\left(\mathbf{F}_{0}\right), \\
& \mathbf{F}_l^{\prime}=[\mathcal{P}\left(\text{GELU}\left(\text{MLP}\left(\mathbf{F}_l\right)\right)\right)]_{\times 2}+\mathcal{P}\left(\text{Sigmoid}\left(\text{MLP}\left(\mathbf{F}_l\right)\right)\right), \\
& \hat{\mathbf{F}_{l}}=\text{Interpolate}\left(\mathbf{F}_l^{\prime}\right) \odot \mathbf{F}_{0},
\end{aligned}
\end{equation}
where $\text{Interpolate}(\cdot)$ and $\mathcal{P}(\cdot)$ denote the interpolation and permute functions. $\text{GELU}(\cdot)$ and $\text{Sigmoid}(\cdot)$ represents GELU and Sigmoid functions. MLP  is a $1 \times 1$ convolution layer. $\odot$ represents element-wise multiplication.

\vspace{-1mm}

{\flushleft\textbf{Feed-forward layer}.}
Similar to vision Transformer (ViT)~\cite{sharir2021image}, FFL performs dimension reduction and non-linear transformations. Here, we adopt a FFL to transform features into compact representations, which is defined as follows:
\begin{equation}
\mathbf{F}_{l}=\text{Conv}_{1 \times 1}\left(\text{GELU}\left(\text{Conv}_{3 \times 3}\left(\left(\mathbf{F}_{0}\right)\right)\right)\right),
\end{equation}
where $\text{Conv}_{3 \times 3}$ is a $3 \times 3$ convolution layer. The initial $3 \times 3$ convolution is used to enhance locality and increase the number of channels for channel mixing. The later $1 \times 1$ convolution is adopted for reducing the channels back to the original input dimension.

\vspace{-1mm}

\subsection{Frequency feature mixing block}
We note that existing methods~\cite{wang2023ultra,li2023embedding,ren2023fast,wu2024mixnet} mostly employ direct downsampling of UHD images to create low-resolution versions, aiming to reduce computational burden. In such cases, the full-resolution restoration process is predominantly governed by information learned solely from the low-resolution images, resulting in suboptimal performance and a tendency to lose high-frequency details, which are abundant in UHD images. To this end, we develop the FFMB as a unit for the auxiliary branch. It leverages the frequency domain information of the full-resolution images and guides the decoding restoration process of the main branch composed of SFMB. Given the input feature $\mathbf{Y}_{l-1} \in \mathbb{R}^{H \times W \times C}$, the proposed FFMB can be formulated as:

\begin{equation}
\begin{aligned}
& \mathbf{Y}_l^{\prime}=\mathbf{Y}_{l-1}+\text{FFML}\left(\text{LN}\left(\mathbf{Y}_{l-1}\right)\right), \\
& \mathbf{Y}_l=\mathbf{Y}_l^{\prime}+\text{FFL}\left(\text{LN}\left(\mathbf{Y}_l^{\prime}\right)\right), 
\end{aligned}
\end{equation}
where $\mathbf{Y}_l^{\prime}$ and $\mathbf{Y}_l$ are output feature from the frequency feature modulation layer (FFML) and FFL. The structure of FFL remains the same as that in SFMB.

\vspace{-1mm}

{\flushleft\textbf{Frequency feature modulation layer}.}
According to the convolution theorem, convolution in one domain is mathematically equivalent to the Hadamard product in its corresponding Fourier domain~\cite{huang2023adaptive}. This also motivates us to introduce a FFML for implementing frequency-space manipulation. Given an input feature $\mathbf{F}_{0}$,  we employ Fast Fourier Transform (FFT) to obtain the corresponding frequency representations. Then, we adopt two stacks of MLP layers with a ReLU layer in between. Finally, we perform an inverse FFT and interact with the input features to obtain updated feature representations in the original latent space. This process can be formulated as:
\begin{equation}
\begin{aligned}
\mathbf{F}_{l} & =\mathcal{F}^{-1}\left(\text{MLP}\left(\text{ReLU}\left(\text{MLP}\left(\mathcal{F}\left(\mathbf{F}_{0}\right)\right)\right)\right)\right), \\
\hat{\mathbf{F}_{l}} & =\mathbf{F}_{l} \odot \mathbf{F}_{0},
\end{aligned}
\end{equation}
where $\mathcal{F}(\cdot)$ denotes the FFT and $\mathcal{F}^{-1}(\cdot)$ denotes the inverse FFT.

\section{Experiments}
\vspace{-2mm}
In this section, we first present the experimental settings of our proposed UDR-Mixer. Then we conduct a benchmark study on our method and other comparative methods. More results can be found in the supplementary material.

\vspace{-1mm}

\subsection{Experimental settings}

{\flushleft\textbf{Implementation details}.}
In our model, $\left\{N_1, N_2, N_3\right\}$ are set to $\{2,2,4\}$. The initial number of feature channels for the main and auxiliary branches is set to 48 and 64, respectively. We conduct model training on four NVIDIA GeForce RTX 3090 GPUs with 24GB memory. In total, we perform 500 epochs of training. During the training, we adopt the Adam optimizer with a learning rate of $2 \times 10^{-4}$. The patch size is set to be $768 \times 768$ pixels and the batch size is set to be 8. To augment the training data, we apply random horizontal and vertical flips. For testing UHD images, we use one NVIDIA TESLA V100 with 32GB memory.

\vspace{-1mm}

{\flushleft\textbf{Compared methods}.}
We compare our MLP-based approach with prior-based algorithms (i.e., DSC~\cite{luo2015removing}), CNN-based networks (i.e., LPNet~\cite{fu2019lightweight}, JORDER-E~\cite{yang2019joint}, RCDNet~\cite{wang2020model}, SPDNet~\cite{yi2021structure}), and Transformer-based models (i.e., IDT~\cite{xiao2022image}, Restormer~\cite{zamir2022restormer}
, DRSformer~\cite{chen2023learning}, and UDR-S2Former~\cite{chen2023sparse}). For fair comparison, we utilize the official released code of these approaches. All deep learning-based methods are retrained on the proposed 4K-Rain13k dataset. We uniformly select the weights from their final training epoch for testing purposes. Note that for some approaches (JORDER-E~\cite{yang2019joint}, RCDNet~\cite{wang2020model}, SPDNet~\cite{yi2021structure}, Restormer~\cite{zamir2022restormer}, and DRSformer~\cite{chen2023learning}), we are unable to infer full-resolution results on UHD images. Following previous UHD studies~\cite{zheng2021ultra,li2023embedding}, we adjust the input to the largest size that the model can handle and then resize the image to the original resolution.

\vspace{-1mm}

{\flushleft\textbf{Evaluation metrics}.}
For the 4K-Rain13k benchmark with ground truth images, we employ full-reference metrics PSNR~\cite{huynh2008scope}, SSIM~\cite{wang2004image} and MSE to evaluate the image quality of each restored results. For the real-world scenes without ground truth images, we adopt the non-reference metrics NIQE~\cite{mittal2012no}, PIQE~\cite{venkatanath2015blind}, and BRISQUE~\cite{mittal2012making}. Higher PSNR and SSIM values signify better restoration quality, while lower MSE, NIQE, PIQE and BRISQUE scores indicate better perceptual quality. We also test the trainable parameters and FLOPs to analyze the computational complexity of the model.

\begin{table*}[t]
	\centering
	\caption{Quantitative evaluations on the proposed 4K-Rain13k dataset. ``Params'' and ``FLOPs'' represent the number of trainable model parameters (in M) and FLOPs (in G), respectively. The results of FLOPs are tested on the images with $1024 \times 1024$ pixels.}
	\resizebox{1.0\textwidth}{!}{
		\begin{tabular}{ll|c|ccc|cc}
			\toprule
			\multicolumn{2}{l|}{\multirow{2}{*}{Methods}}                                  & \multirow{3}{*}{Venue} & \multicolumn{3}{c|}{4K-Rain13k}                    & \multicolumn{2}{c}{Complexity} \\ \cline{4-8} 
			\multicolumn{2}{l|}{}                                                          &                        & PSNR $\uparrow$          & SSIM $\uparrow$           & MSE $\downarrow$          & Params       & FLOPs       \\ \cline{1-2} \cline{4-8} 
			\multicolumn{2}{l|}{Rainy input}                                               &                        & 21.14          & 0.7594          & 812.98          & -              & -             \\ \hline
			\multicolumn{1}{l|}{Prior-based methods}                        & DSC~\cite{luo2015removing}          & ICCV'15                & 22.93          & 0.6299          & 498.63          & -              & -             \\ \hline
			\multicolumn{1}{l|}{\multirow{4}{*}{CNN-based methods}}         & LPNet~\cite{fu2019lightweight}        & TNNLS'19                & 27.86          & 0.8924          & 171.33          & 0.03           & 57.1          \\
			\multicolumn{1}{l|}{}                                           & JORDER-E~\cite{yang2019joint}     & TPAMI'19                & 30.46          & 0.9117          & 103.93          & 4.21           & 4363.3          \\
			\multicolumn{1}{l|}{}                                           & RCDNet~\cite{wang2020model}       & CVPR'20                & 30.83          & 0.9212          & 95.40          & 3.17           & 3400.3          \\
			\multicolumn{1}{l|}{}                                           & SPDNet~\cite{yi2021structure}       & ICCV'21                & 31.81          & 0.9223          & 78.17          & 3.04           & 1428.8          \\ \hline
			\multicolumn{1}{l|}{\multirow{4}{*}{Transformer-based methods}} & IDT~\cite{xiao2022image}          & TPAMI'22                & 32.91          & 0.9479          & 57.04         & 16.41           & -          \\
			\multicolumn{1}{l|}{}                                           & Restormer~\cite{zamir2022restormer}    & CVPR'22                & 33.02          & 0.9335          & 60.43          & 26.12           & 2478.1          \\
			\multicolumn{1}{l|}{}                                           & DRSformer~\cite{chen2023learning}    & CVPR'23                & 32.96          & 0.9334          & 62.95          & 33.65           & 3887.8          \\
			\multicolumn{1}{l|}{}                                           & UDR-S2Former~\cite{chen2023sparse} & ICCV'23                & 33.36          & 0.9458          & 50.69          & 8.53           & 395.8          \\ \hline
			\multicolumn{1}{l|}{\multirow{1}{*}{MLP-based method}}        & UDR-Mixer   & Ours                   & \textbf{34.30} & \textbf{0.9505} & \textbf{42.03} & 4.90           & 200.1          \\ \bottomrule
		\end{tabular}
	}
	\label{table2}
 \vspace{-2mm}
\end{table*}

\begin{figure}[!t]
	\centering 	
	\begin{subfigure}[t]{0.19\columnwidth}
		\centering
		\includegraphics[width=\columnwidth]{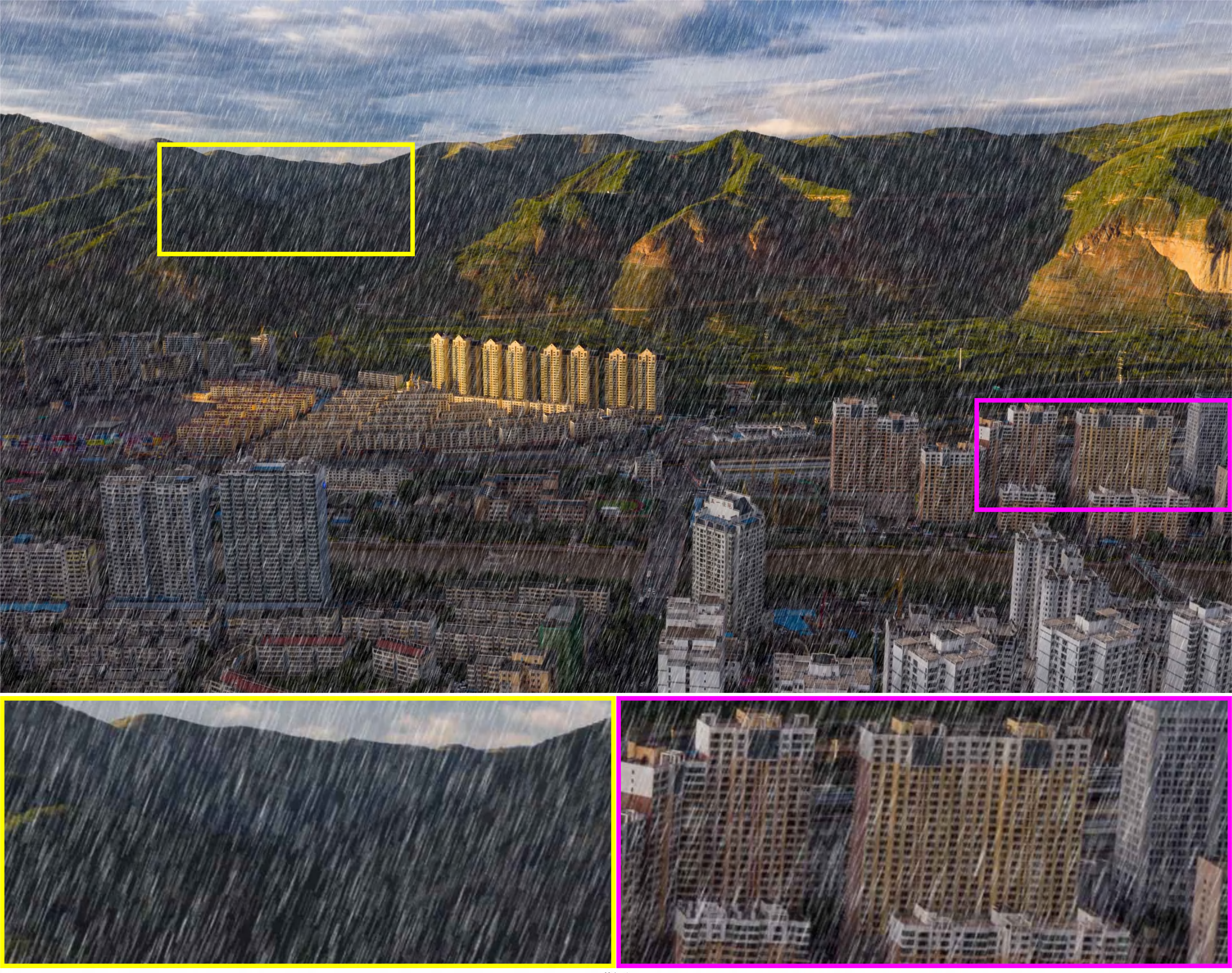}
		\caption{Rainy input}
	\end{subfigure}
	\begin{subfigure}[t]{0.19\columnwidth}
		\centering
		\includegraphics[width=\columnwidth]{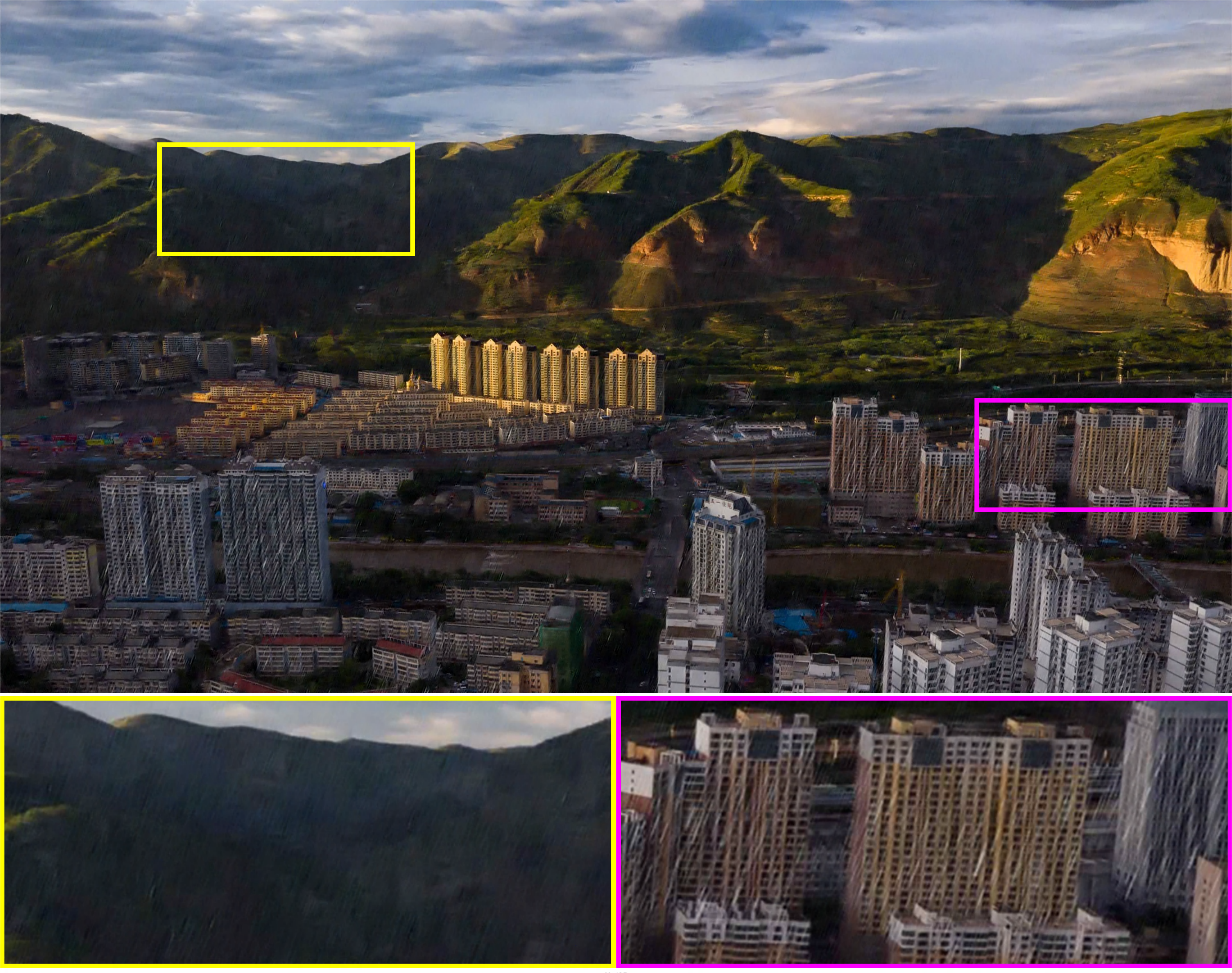}
		\caption{JORDER-E~\cite{yang2019joint}}
	\end{subfigure}
	\begin{subfigure}[t]{0.19\columnwidth}
		\centering
		\includegraphics[width=\columnwidth]{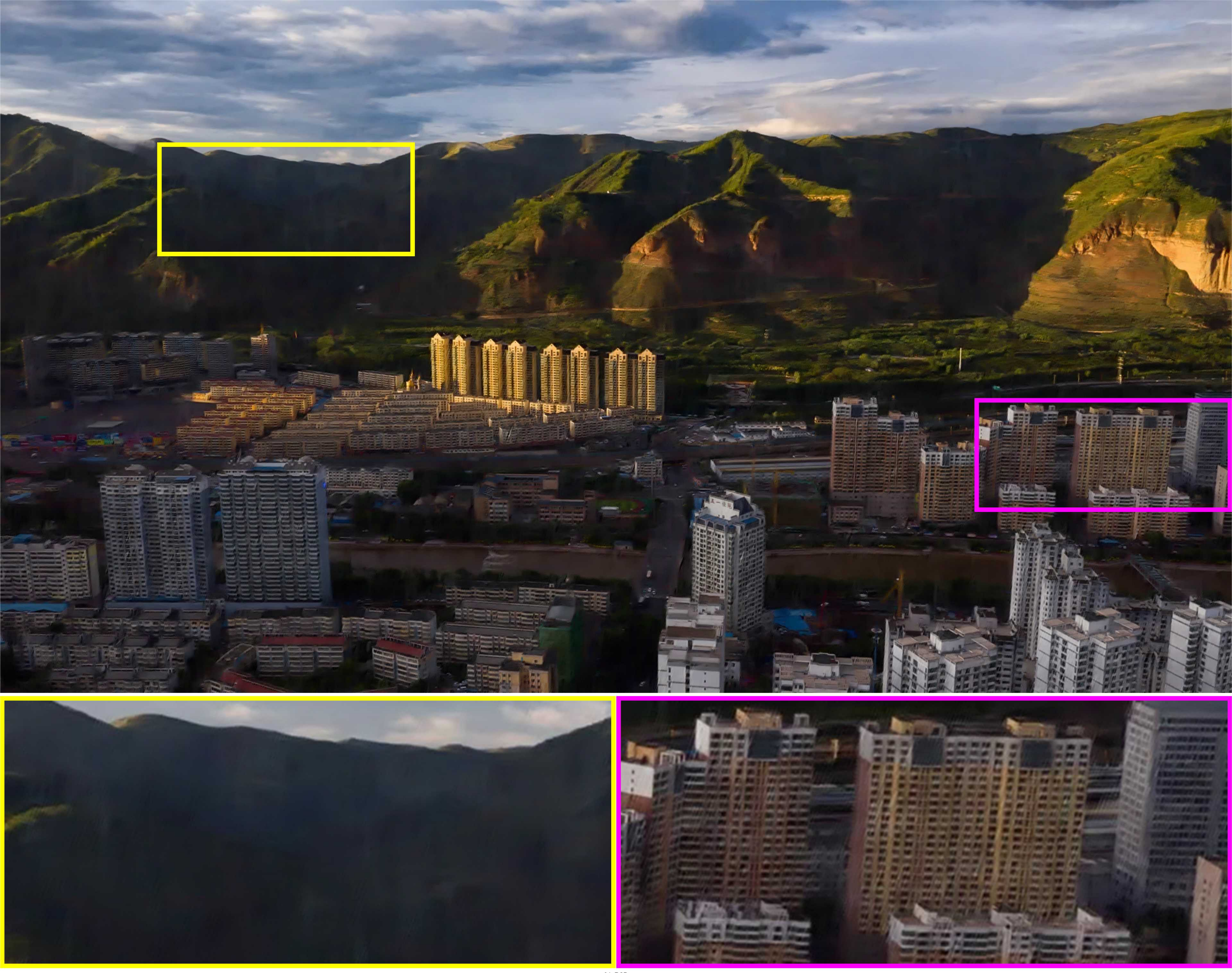}
		\caption{RCDNet~\cite{wang2020model}}
	\end{subfigure}
	\begin{subfigure}[t]{0.19\columnwidth}
		\centering
		\includegraphics[width=\columnwidth]{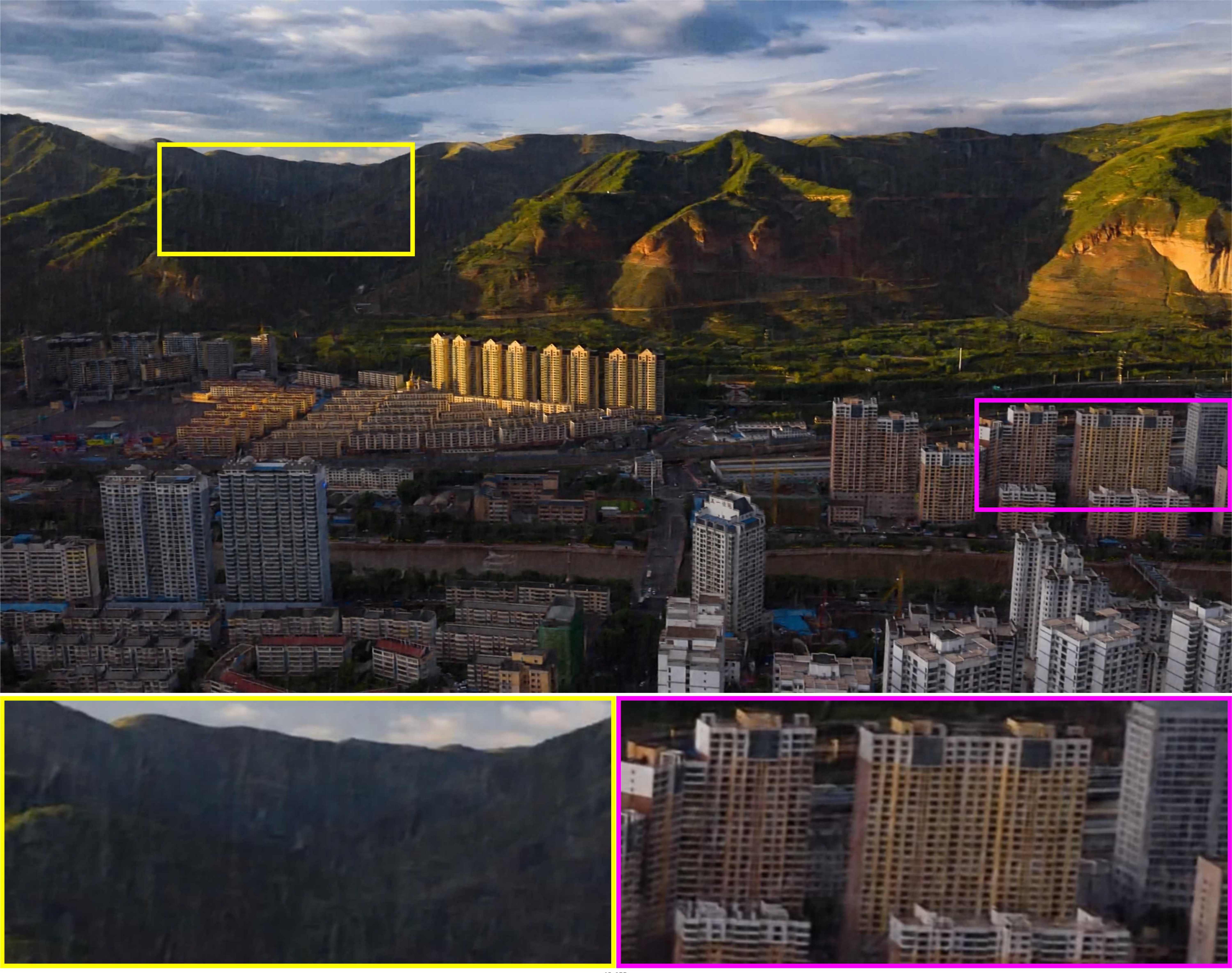}
		\caption{SPDNet~\cite{yi2021structure}}
	\end{subfigure}
	\begin{subfigure}[t]{0.19\columnwidth}
		\centering
		\includegraphics[width=\columnwidth]{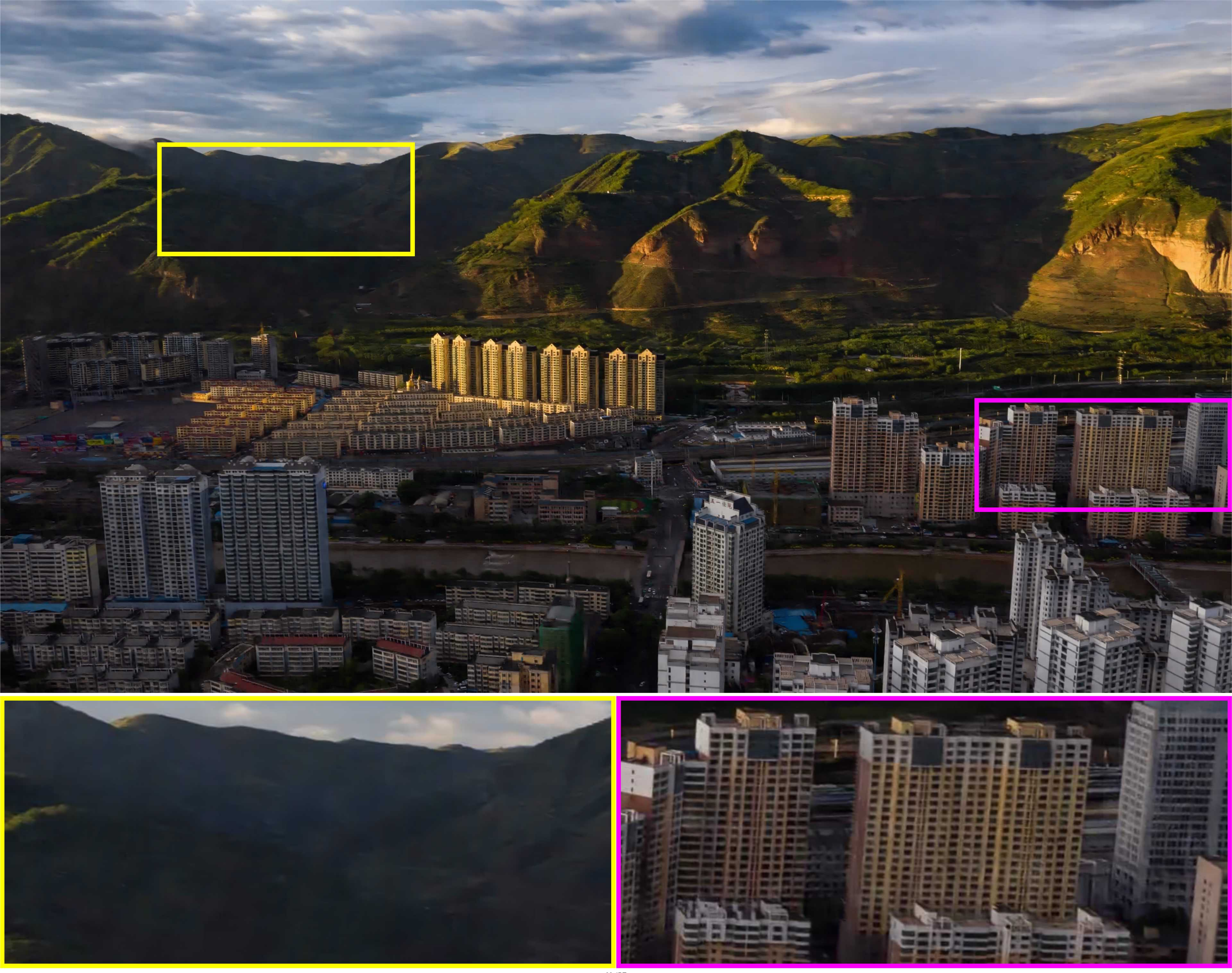}
		\caption{IDT~\cite{xiao2022image}}
	\end{subfigure}
 	\\
	\begin{subfigure}[t]{0.19\columnwidth}
		\centering
		\includegraphics[width=\columnwidth]{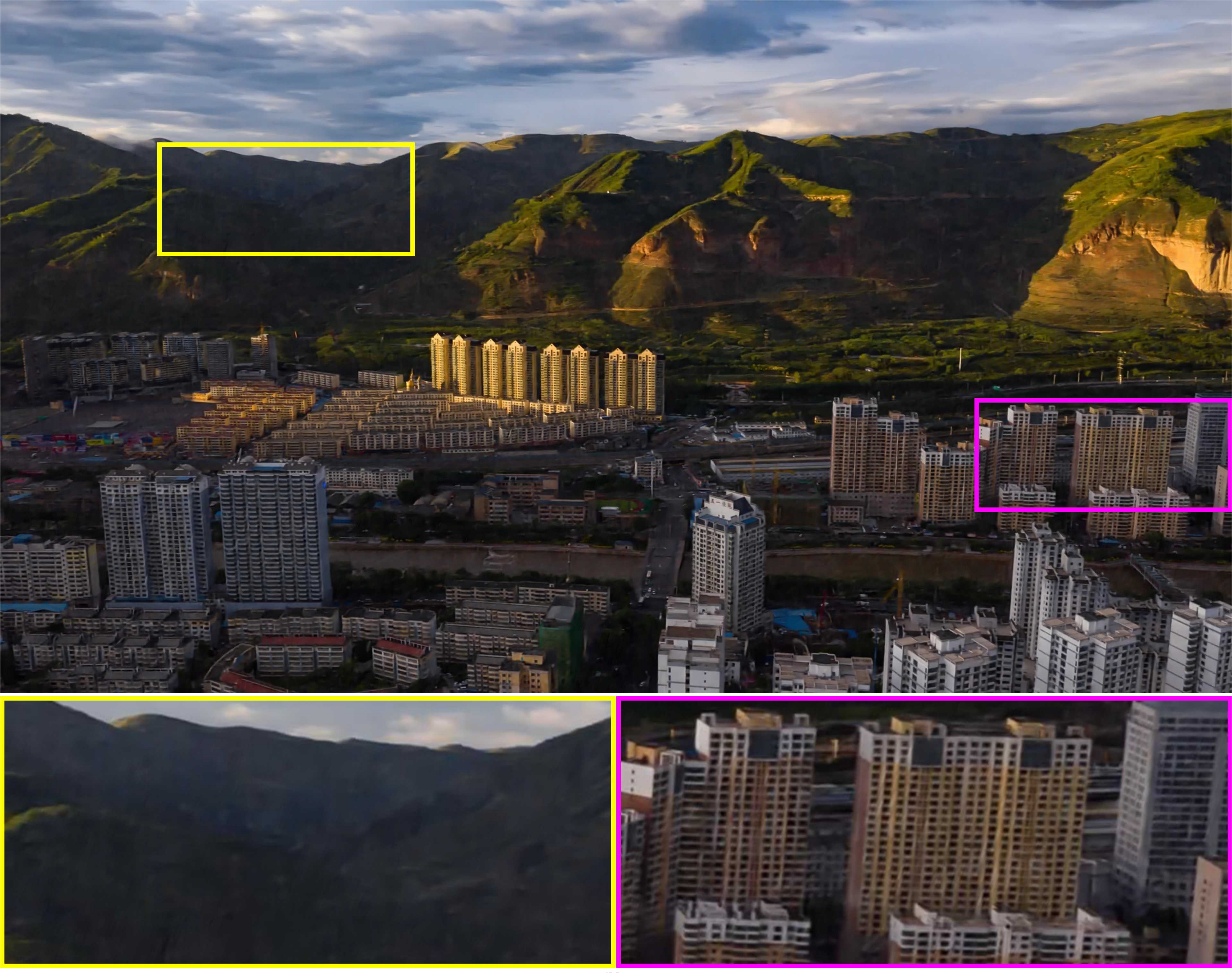}
		\caption{Restormer~\cite{zamir2022restormer}}
	\end{subfigure}
	\begin{subfigure}[t]{0.19\columnwidth}
		\centering
		\includegraphics[width=\columnwidth]{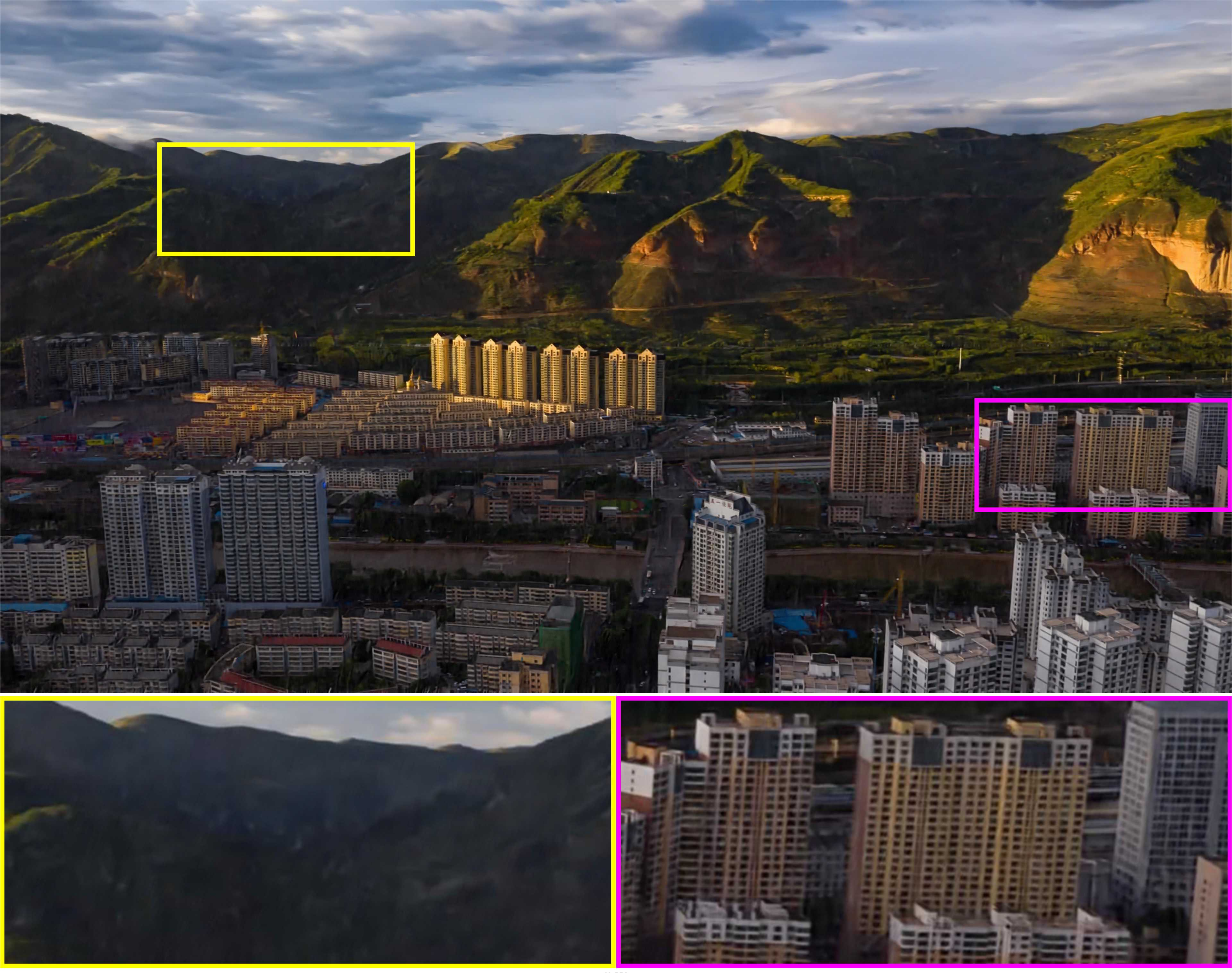}
		\caption{DRSformer~\cite{chen2023learning}}
	\end{subfigure}
	\begin{subfigure}[t]{0.19\columnwidth}
		\centering
		\includegraphics[width=\columnwidth]{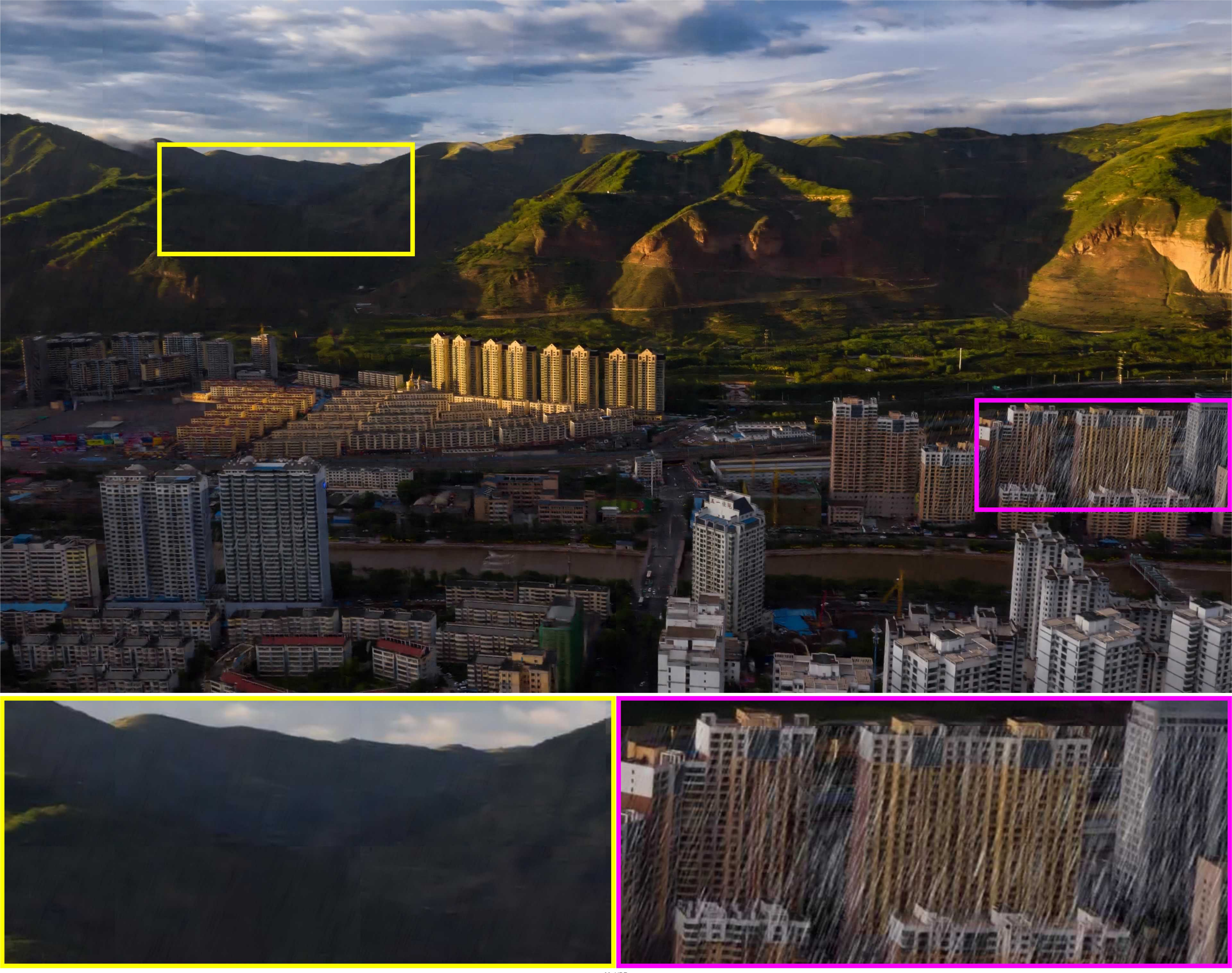}
		\caption*{\raggedright \scriptsize(h) UDR-S2Former~\cite{chen2023sparse}}
	\end{subfigure}
	\begin{subfigure}[t]{0.19\columnwidth}
		\centering
		\includegraphics[width=\columnwidth]{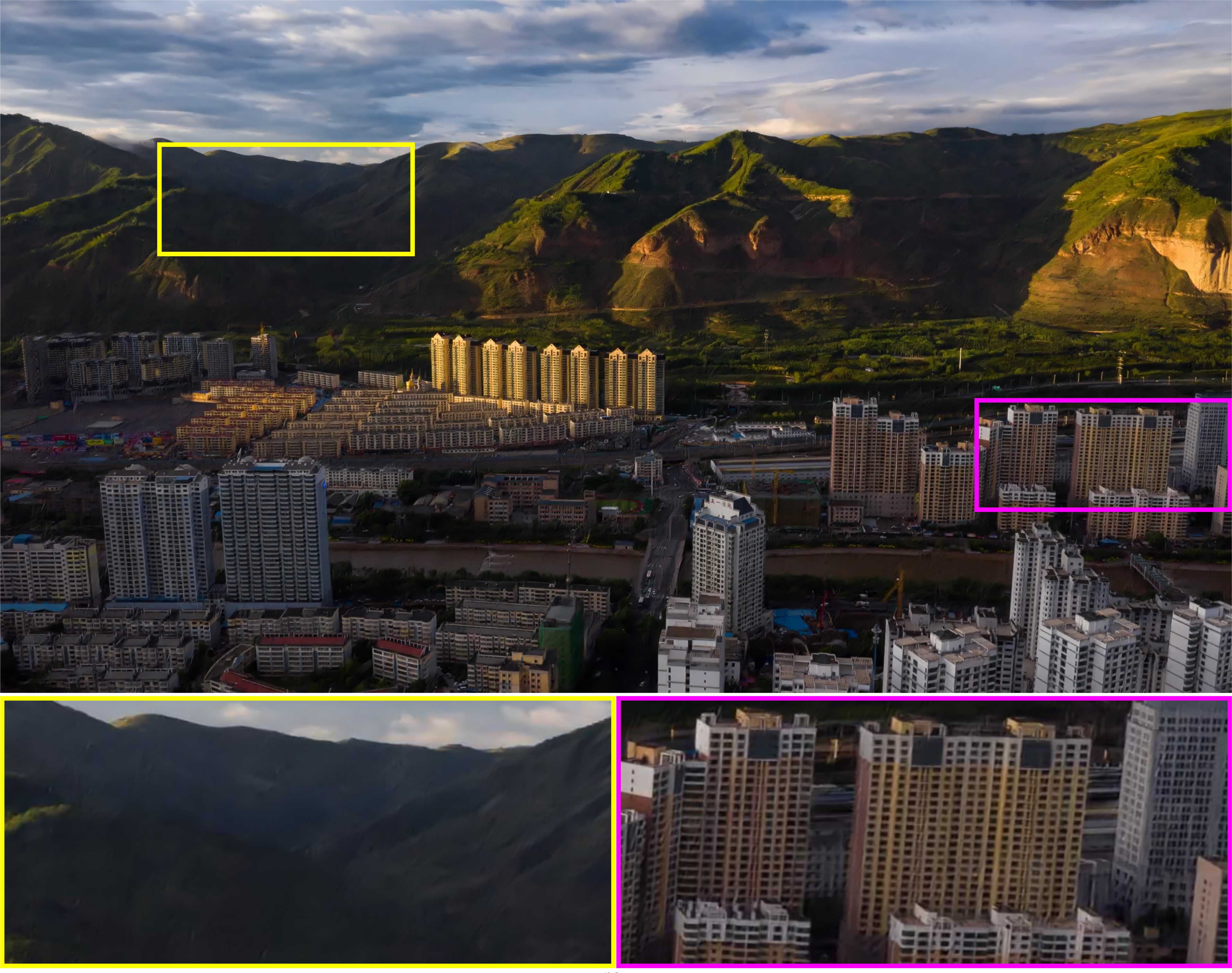}
		\caption*{(i) Ours}
	\end{subfigure}
	\begin{subfigure}[t]{0.19\columnwidth}
		\centering
		\includegraphics[width=\columnwidth]{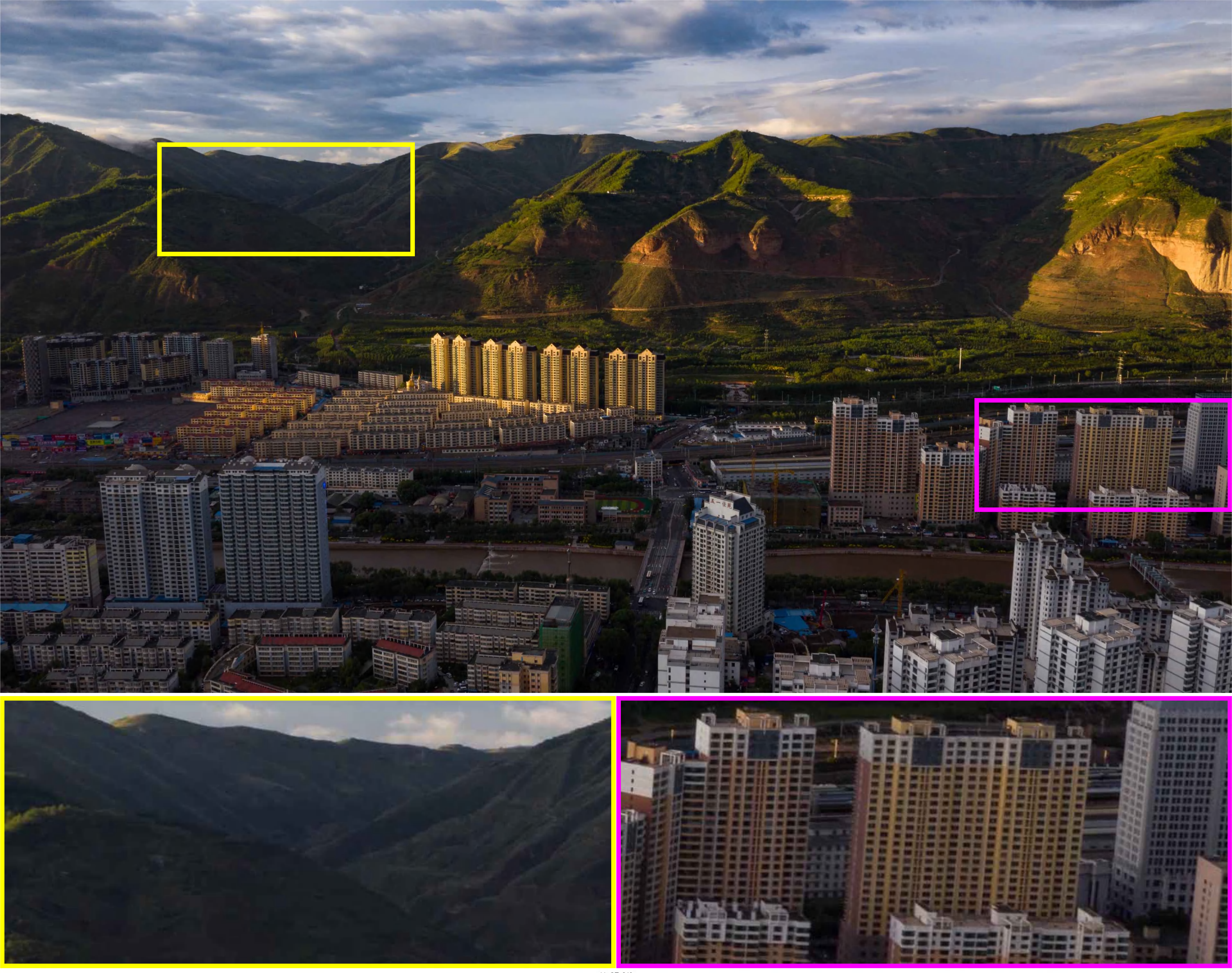}
		\caption*{(j) Ground truth}
	\end{subfigure}
	\caption{Visual quality comparison on the 4K-Rain13k dataset. Compared with the derained results in (b)-(h), our method recovers a high-quality image with clearer details. } 
	\vspace{-4mm}
	\label{fig4} 
\end{figure}

\vspace{-2mm}

\subsection{Comparisons with the state of the art}

{\flushleft\textbf{Evaluations on the proposed 4K-Rain13k}.}
Table \ref{table2} presents the quantitative results of different approaches on the proposed 4K-Rain13k dataset. It can be observed that our proposed UDR-Mixer achieves the highest PSNR and SSIM values while maintaining the lowest MSE value, indicating the superiority of our method in rain removal from UHD images. Specifically, our method outperforms the state-of-the-art UDR-S2Former~\cite{chen2023sparse} by 0.94dB in terms of PSNR, while utilizing fewer network parameters and lower FLOPs. 
In Figure \ref{fig4}, we further present the visual results of different methods. We observe that CNN-based methods struggle to recover texture details, such as those in building areas, under the influence of densely packed rain streaks. Additionally, despite its ability to model global information, UDR-S2Former~\cite{chen2023sparse}, as a Transformer-based method, still exhibits sensitivity to spatially-long rain streaks present in UHD images, resulting in residual rain artifacts. In contrast, our approach produces clearer images while preserving high-frequency information. 

\vspace{-1mm}

\begin{figure}[!t]
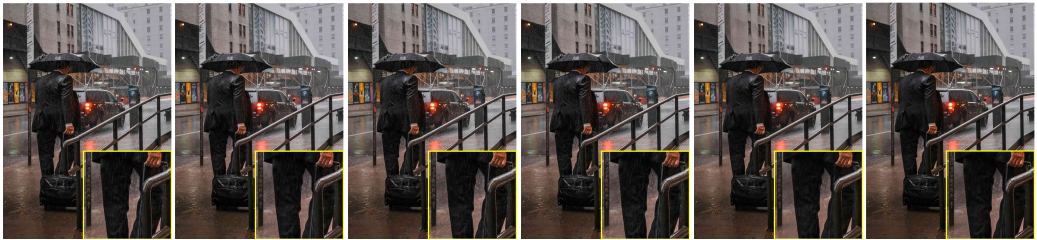
\scriptsize
	\centering 	
	\begin{subfigure}[t]{0.16\columnwidth}
		\centering
		\includegraphics[width=\columnwidth]{figures/Figure5/a_Input.pdf}
		\caption{Rainy input}
	\end{subfigure}
	\begin{subfigure}[t]{0.16\columnwidth}
		\centering
		\includegraphics[width=\columnwidth]{figures/Figure5/b_SPDNet.pdf}
		\caption{SPDNet}
	\end{subfigure}
	\begin{subfigure}[t]{0.16\columnwidth}
		\centering
		\includegraphics[width=\columnwidth]{figures/Figure5/c_Restormer.pdf}
		\caption{Restormer}
	\end{subfigure}
	\begin{subfigure}[t]{0.16\columnwidth}
		\centering
		\includegraphics[width=\columnwidth]{figures/Figure5/d_DRSformer.pdf}
		\caption{DRSformer}
	\end{subfigure}
	\begin{subfigure}[t]{0.16\columnwidth}
		\centering
		\includegraphics[width=\columnwidth]{figures/Figure5/e_UDR-S2Former.pdf}
		\caption*{\raggedright \scriptsize(e) UDR-S2Former}
	\end{subfigure}
	\begin{subfigure}[t]{0.16\columnwidth}
		\centering
		\includegraphics[width=\columnwidth]{figures/Figure5/f_Ours.pdf}
		\caption*{(f) Ours}
	\end{subfigure}
	\caption{Visual quality comparison on a real-world UHD rainy image from the collected 4K-RealRain. Compared with the derained results in (b)-(e), our method recovers a clearer image.} 
	\vspace{-5mm}
	\label{fig5} 
\end{figure}

{\flushleft\textbf{Evaluations on real 4K rainy images}.}
To further evaluate the generalization capability of various deraining methods in real rainy scenes, we collect 320 real 4K rainy images from the Internet and real-world sources, referred to as 4K-RealRain. These scenes mostly originate from high-definition captures using smartphones. The quantitative results for different methods are reported in Table~\ref{table3}. Clearly, our method achieves the lowest values across three metrics: NIQE, PIQE, and BRISQUE. This indicates that, compared to other models, the output results from our UDR-Mixer exhibit clearer content and better perceptual quality in real rainy scenes. Figure \ref{fig5} displays a comparison of visual results. Our method effectively removes most rain streaks and exhibits visually pleasing restoration effects, indicating its capability to generalize well to unseen real-world data types.

\begin{table*}[t]
	\centering
	\caption{Quantitative evaluations on real rainy images. For all methods, we use the pre-trained models from the 4K-Rain13k dataset to evaluate the image deraining capabilities in real-world scenarios.}
	\resizebox{1.0\textwidth}{!}{
		\begin{tabular}{ccccccccc}
			\toprule
			Methods & Input  & RCDNet & SPDNet & IDT    & Restormer & DRSformer & UDR-S2Former & Ours            \\ \hline
			NIQE $\downarrow$   & 8.208  & 9.997  & 9.917  & 9.067  & 8.636     & 8.493     & 8.104        & \textbf{7.509}  \\
			PIQE $\downarrow$   & 54.863 & 63.816 & 64.774 & 55.049 & 60.335    & 60.441    & 55.204       & \textbf{53.104} \\
			BRISQUE $\downarrow$ & 67.855 & 71.967 & 67.461 & 67.100 & 65.102    & 63.823    & 65.177       & \textbf{53.192} \\ \bottomrule
		\end{tabular}
	}
	\label{table3}
 \vspace{-3mm}
\end{table*}

\begin{table*}[t]
	\centering
	\caption{Quantitative evaluations on the RainDS-RS dataset, which contains Syn-RS and Real-RS.}
	\resizebox{1.0\textwidth}{!}{
		\begin{tabular}{ccccccccc}
			\toprule
			Methods & JORDER-E  & MSPFN & MPRNet & Uformer    & Restormer & IDT & UDR-S2Former & Ours            \\ \hline
			PSNR $\uparrow$  & 30.11  & 32.53  & 34.05  & 33.79  & 34.41     & 34.56     & 35.15        & \textbf{35.72}  \\
			SSIM $\uparrow$ & 0.819 & 0.851 & 0.859 & 0.851 & 0.861    & 0.863    & 0.867       & \textbf{0.868} \\ \bottomrule
		\end{tabular}
	}
	\label{table4}
 \vspace{-3mm}
\end{table*}

\vspace{-2mm}

{\flushleft\textbf{Evaluations on low-resolution benchmarks}.}
We further validate the scalability of our method on low-resolution benchmarks. Following~\cite{chen2023sparse}, we conduct experiments on the RainDS-RS dataset~\cite{quan2021removing}, which contains Syn-RS and Real-RS subnets. Here, we adjust our UDR-Mixer model for a fair comparison. Specifically, we remove the pixel unshuffle/shuffle operations used for UHD images in the main branch, while keeping other components consistent. For distinction, we name it UDR-Mixer-L. According to the quantitative results in Table \ref{table4}, our proposed method not only demonstrates satisfactory deraining effects on 4K images but also proves effective in low-resolution scenes.

\vspace{-2mm}

\subsection{Ablation study}

{\flushleft\textbf{Effectiveness of rearrange strategy in SFMB}.}
We first replace the proposed SFMBs with residual blocks
that have comparable parameters as the baseline model (i). Table \ref{table5} shows that our method improves restoration performance better compared to the baseline model by introducing SFMB. The feature rearrangement strategy is a critical component of our proposed SFMB. Here, we compare with recent MLP-based feature rearrangement methods, including spatial shift~\cite{yu2022s2}, height-direction region rearrange and width-direction region rearrange~\cite{guo2022hire}. Compared to methods (ii-iv), our dimension rearrangement approach yields superior quantitative results. The visual results in Figure \ref{fig7} (b-d) and (f) also demonstrate that our method not only effectively removes complex rain streaks but also better preserves the fine details of the image. The reason behind this lies in our method implicitly enhancing the capture of multi-view features through dimension transformation, making it more suitable for modeling long-range spatial relationships in UHD images.

\vspace{-1mm}
	
{\flushleft\textbf{Effect of the number of permute stages}.}
We further analyze the effect of the number of permute stages in the SFMB. Note that we utilize Permute operations to rotate rotate 3D feature maps between adjacent stages. When stage=1, the model (v) can only capture the single-view features. We find that through multiple stages of recursive encoding, features learned from three perspectives are effectively correlated, thus aiding in further boosting the image restoration performance.

\vspace{-1mm}

{\flushleft\textbf{Effectiveness of FFMB}.}
The FFMB in the auxiliary branch is used to better explore frequency information in our UDR-Mixer for high-quality UHD image restoration. To demonstrate the effectiveness of this branch, we remove this component and investigate its influence in Table \ref{table5}. In comparison to our approach, the restoration performance of model (vii) is suboptimal. In addition, Figures \ref{fig7} (e) and (f) also show that our method generates much clearer details. 

\begin{table*}[t]\footnotesize
	\centering
 \vspace{-2mm}
	\caption{Ablation comparison on different variants of our UDR-Mixer on the 4K-Rain13k dataset.}
	\resizebox{1.0\textwidth}{!}{
		\begin{tabular}{c|cccccc|c|cc}
			\toprule
			\multirow{3}{*}{Methods} & \multicolumn{6}{c|}{SFMB}                                                                                                                                                                                                                                                                                                                          & \multirow{3}{*}{FFMB} & \multirow{3}{*}{PSNR} & \multirow{3}{*}{SSIM} \\ \cline{2-7}
			& \multicolumn{1}{c|}{\multirow{2}{*}{\begin{tabular}[c]{@{}c@{}}Spatial\\ Shift\end{tabular}}} & \multicolumn{1}{c|}{\multirow{2}{*}{\begin{tabular}[c]{@{}c@{}}H-Region\\ Rearrange\end{tabular}}} & \multicolumn{1}{c|}{\multirow{2}{*}{\begin{tabular}[c]{@{}c@{}}W-Region\\ Rearrange\end{tabular}}} & \multicolumn{3}{c|}{Dimension Rearrange} &                       &                       &                       \\
			& \multicolumn{1}{c|}{}                                                                         & \multicolumn{1}{c|}{}                                                                              & \multicolumn{1}{c|}{}                                                                              & Stage-1      & Stage-2     & Stage-3     &                       &                       &                       \\ \hline
			(i)                      & \multicolumn{1}{c|}{}                                                                         & \multicolumn{1}{c|}{}                                                                              & \multicolumn{1}{c|}{}                                                                              &              &             &             &                       & 33.41                 & 0.9417                \\
			(ii)                      & \multicolumn{1}{c|}{\CheckmarkBold}                                                                        & \multicolumn{1}{c|}{}                                                                              & \multicolumn{1}{c|}{}                                                                              &              &             &             &                       & 33.65                 & 0.9433                \\
			(iii)                      & \multicolumn{1}{c|}{}                                                                         & \multicolumn{1}{c|}{\CheckmarkBold}                                                                             & \multicolumn{1}{c|}{}                                                                              &              &             &             &                       & 34.09                 & 0.9437                \\
			(iv)                      & \multicolumn{1}{c|}{}                                                                         & \multicolumn{1}{c|}{}                                                                              & \multicolumn{1}{c|}{\CheckmarkBold}                                                                             &              &             &             &                       & 33.74                 & 0.9413                \\
			(v)                      & \multicolumn{1}{c|}{}                                                                         & \multicolumn{1}{c|}{}                                                                              & \multicolumn{1}{c|}{}                                                                              & \CheckmarkBold            &             &             &                       & 32.41                 & 0.9394                \\
			(vi)                      & \multicolumn{1}{c|}{}                                                                         & \multicolumn{1}{c|}{}                                                                              & \multicolumn{1}{c|}{}                                                                              & \CheckmarkBold            & \CheckmarkBold           &             &                       & 32.50                 & 0.9409                \\
			(vii)                      & \multicolumn{1}{c|}{}                                                                         & \multicolumn{1}{c|}{}                                                                              & \multicolumn{1}{c|}{}                                                                              & \CheckmarkBold            & \CheckmarkBold           & \CheckmarkBold           &                       & 34.15                 & 0.9465                \\
			(viii, Ours)                      & \multicolumn{1}{c|}{}                                                                         & \multicolumn{1}{c|}{}                                                                              & \multicolumn{1}{c|}{}                                                                              & \CheckmarkBold            & \CheckmarkBold           & \CheckmarkBold           & \CheckmarkBold                     & \textbf{34.30}        & \textbf{0.9505}       \\ \bottomrule
		\end{tabular}
	}
	\label{table5}
 \vspace{-3mm}
\end{table*}

\begin{figure*}[t]
	\begin{minipage}{0.4\linewidth}
		\centering
\captionof{table}{Comparisons of model complexity with MAXIM~\cite{tu2022maxim}. The size of the test image is $256 \times 256$.}
		\scalebox{0.9}{
			    \begin{tabular}{l|c|c}
        \toprule
            Methods             & MAXIM~\cite{tu2022maxim}      & UDR-Mixer-L       \\ \hline 
                Params          & 14.1M  & 2.6M\\
                FLOPs          & 216G  & 65G\\
                \bottomrule
    \end{tabular}
		}
		\label{table6}
	\end{minipage}
	\hfill
	\begin{minipage}{0.55\linewidth}
		\centering
		\includegraphics[width=1.0\linewidth]{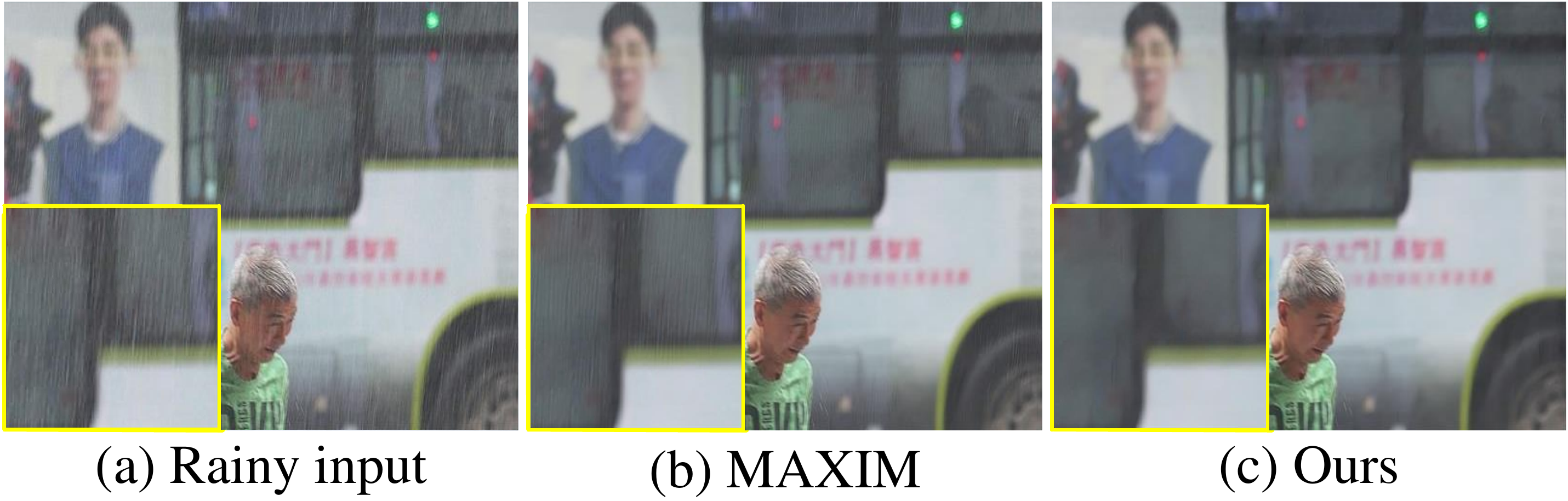}
  \vspace{-6mm}
		\caption{Comparison results with MAXIM~\cite{tu2022maxim}.}
		\label{fig6}
	\end{minipage}
 \vspace{-4mm}
\end{figure*}

\begin{figure}[!t]\scriptsize
	\centering 	
	\begin{subfigure}[t]{0.16\columnwidth}
		\centering
		\includegraphics[width=\columnwidth]{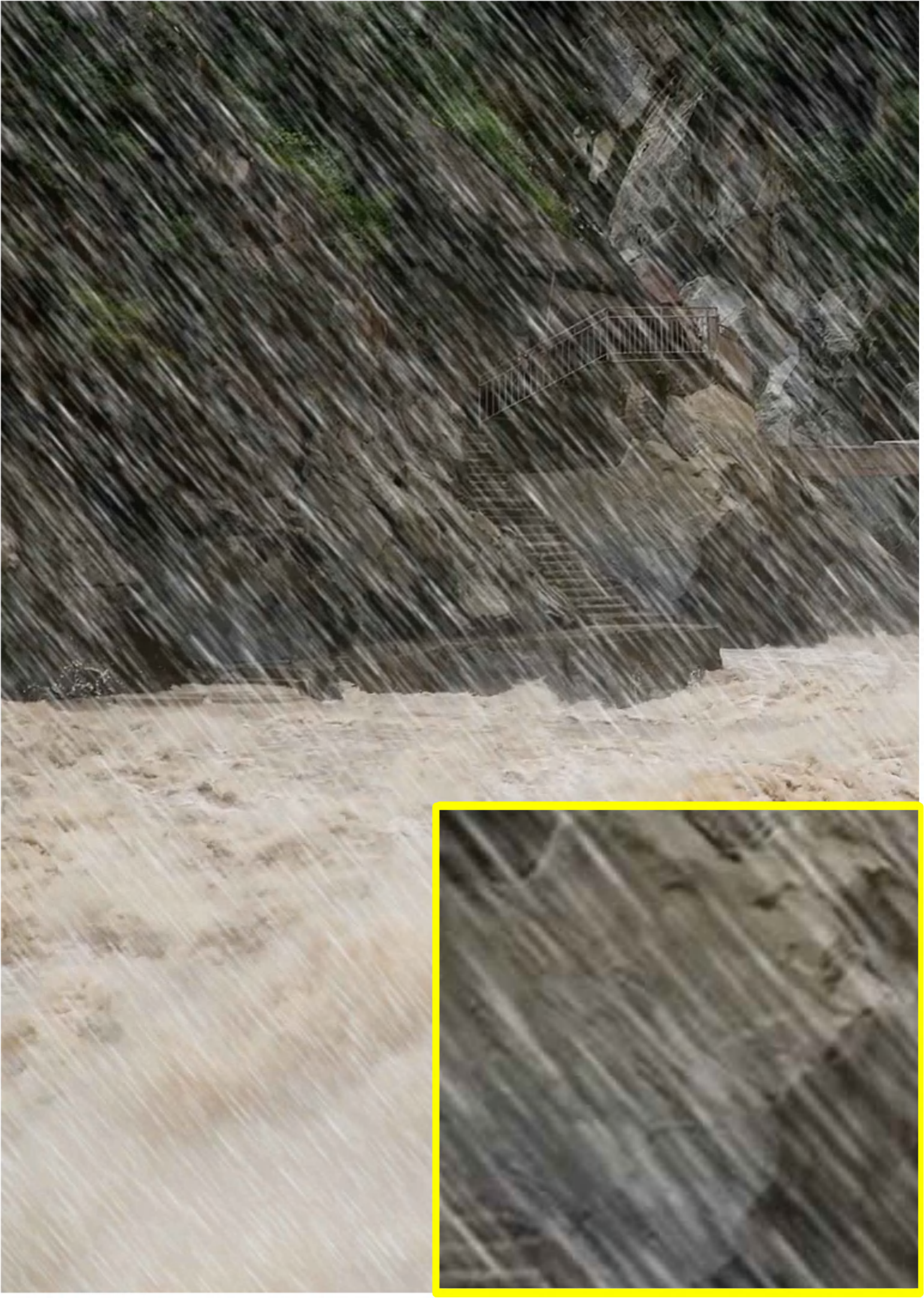}
		\caption{Rainy input}
	\end{subfigure}
	\begin{subfigure}[t]{0.16\columnwidth}
		\centering
		\includegraphics[width=\columnwidth]{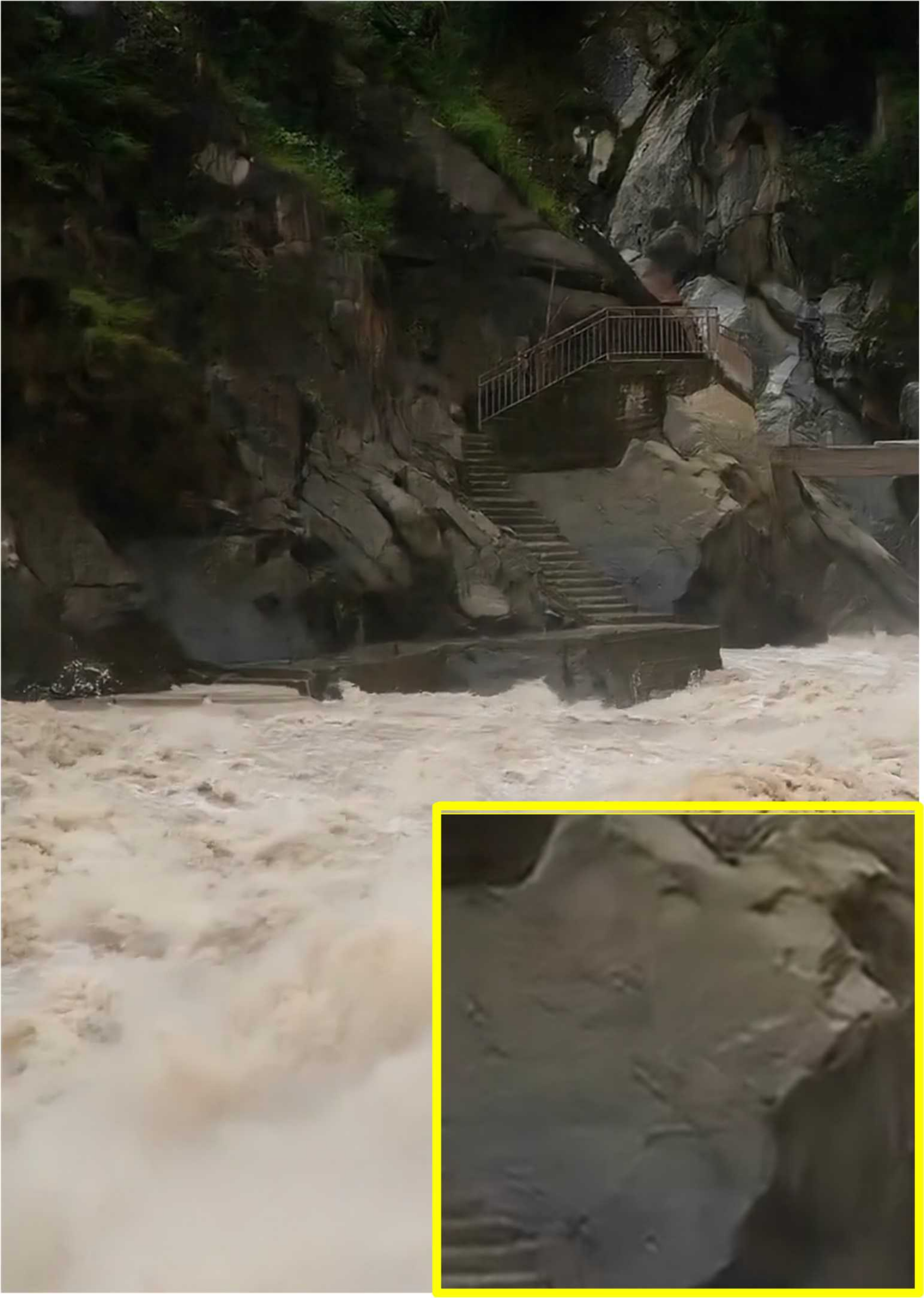}
		\caption{Spatial shift}
	\end{subfigure}
	\begin{subfigure}[t]{0.16\columnwidth}
		\centering
		\includegraphics[width=\columnwidth]{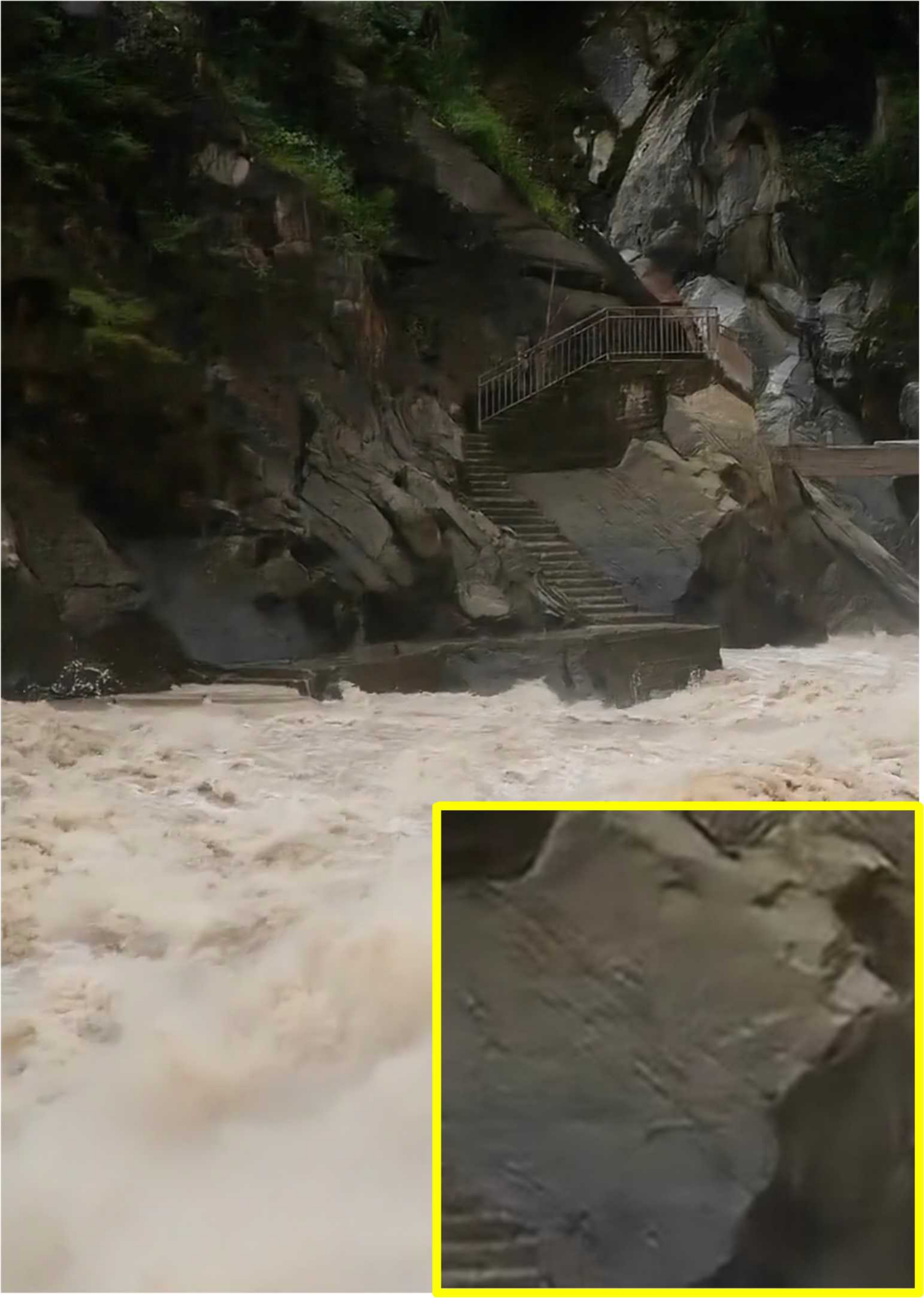}
		\caption{H-region RR}
	\end{subfigure}
	\begin{subfigure}[t]{0.16\columnwidth}
		\centering
		\includegraphics[width=\columnwidth]{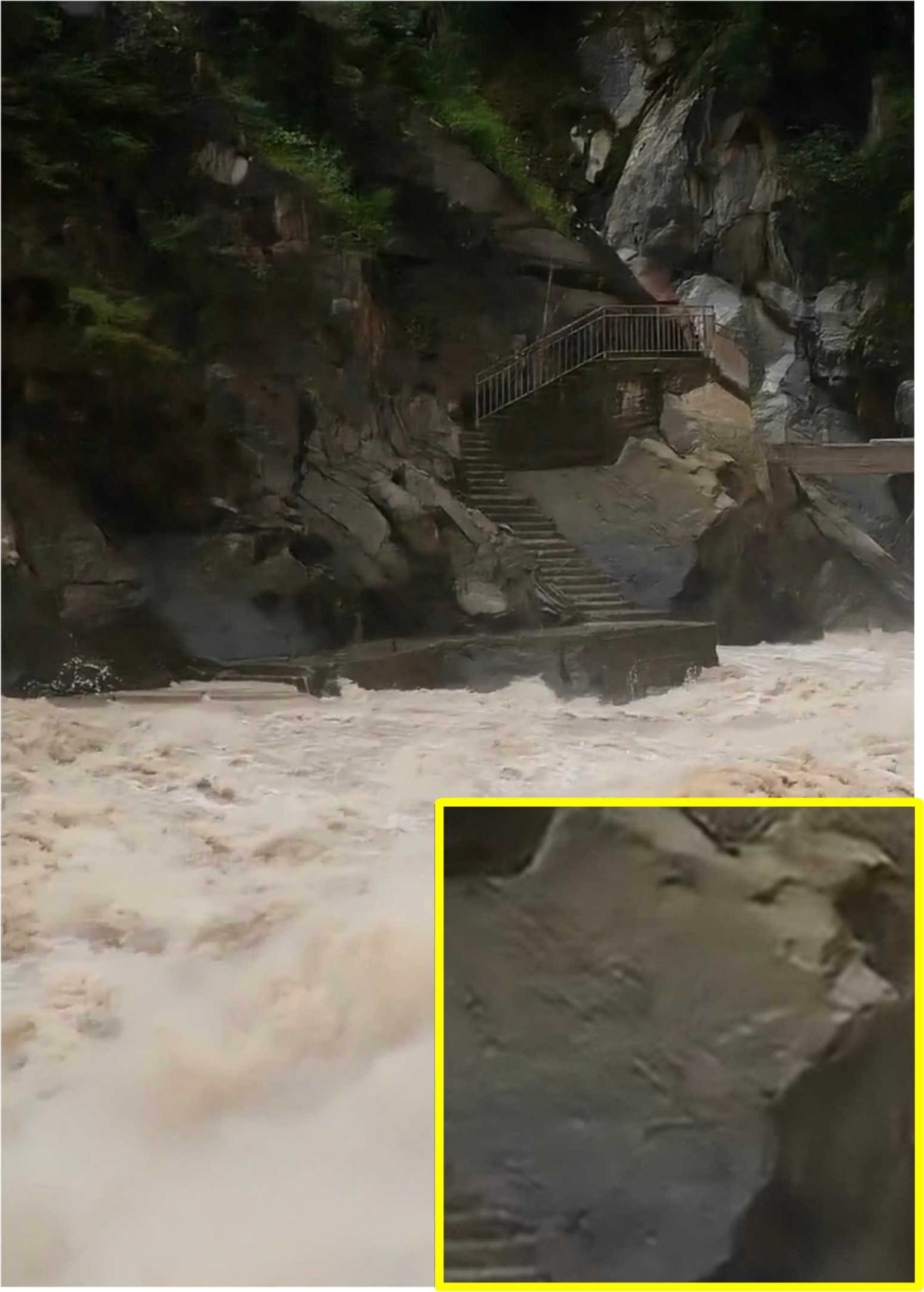}
		\caption{W-region RR}
	\end{subfigure}
	\begin{subfigure}[t]{0.16\columnwidth}
		\centering
		\includegraphics[width=\columnwidth]{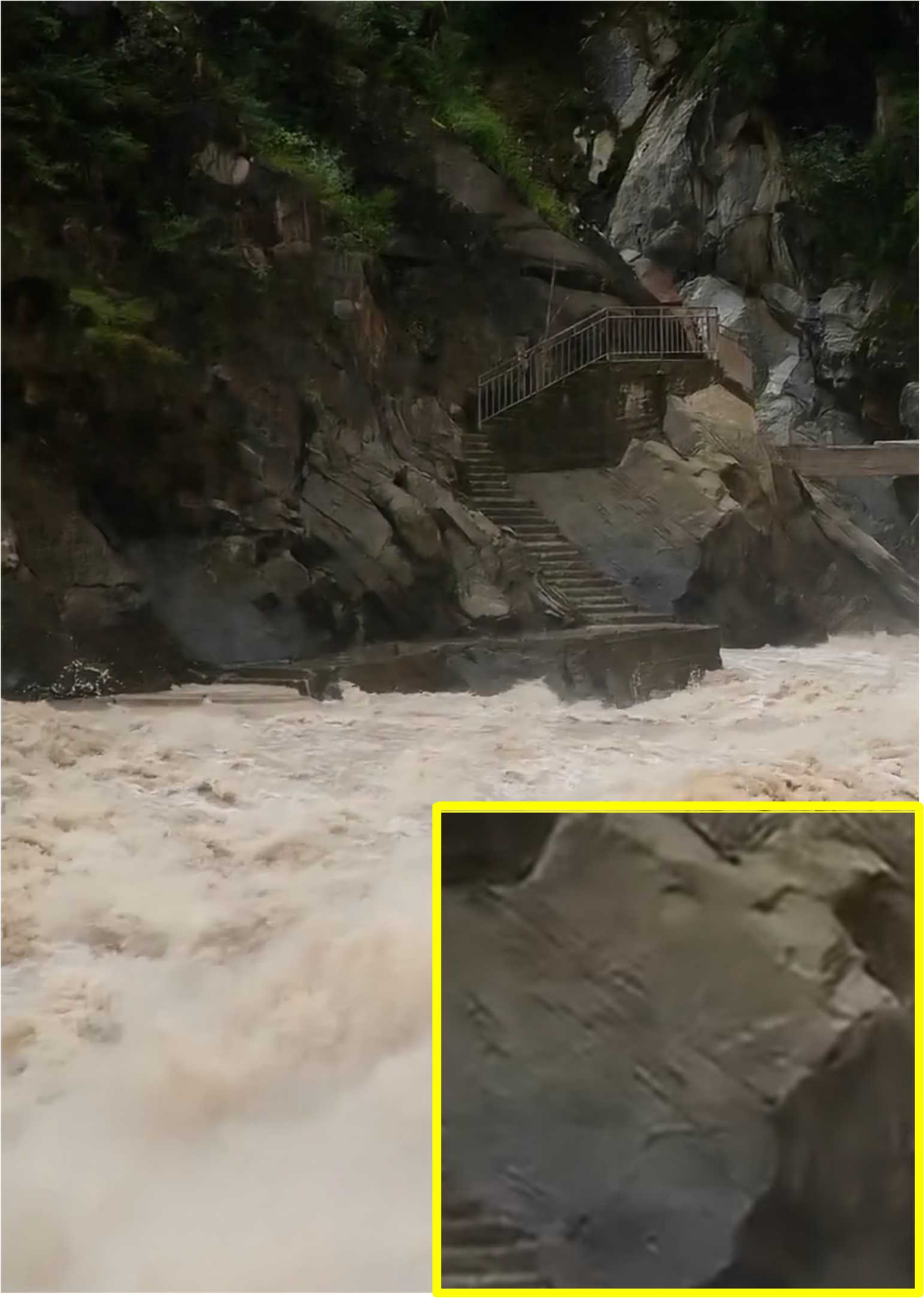}
		\caption{w/o FFMB}
	\end{subfigure}
	\begin{subfigure}[t]{0.16\columnwidth}
		\centering
		\includegraphics[width=\columnwidth]{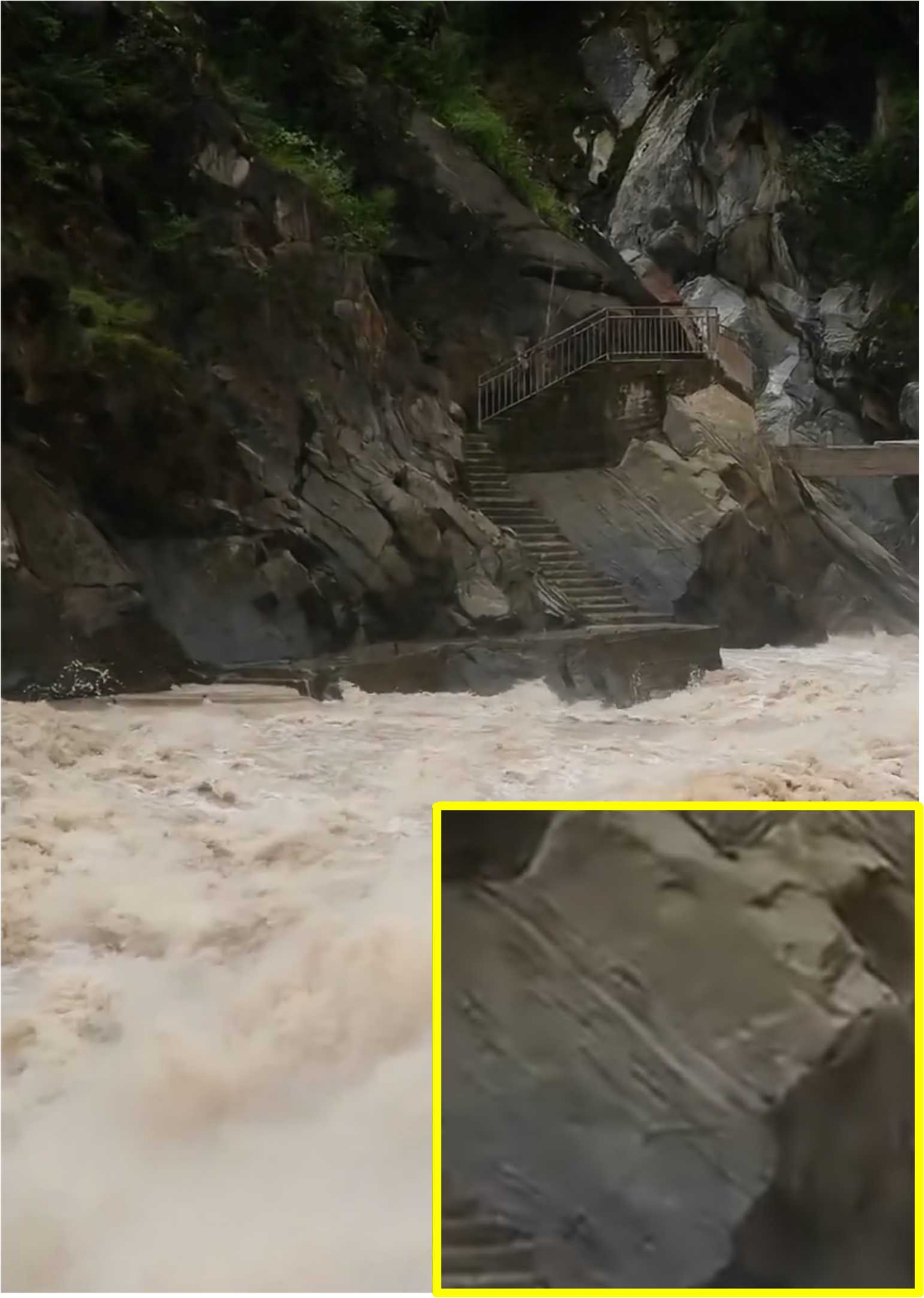}
		\caption{Ours}
	\end{subfigure}
	\caption{Visual comparison on the rearrange strategy in SFMB (b-d) and the proposed FFMB (e).} 
	\vspace{-5mm}
	\label{fig7} 
\end{figure} 

\vspace{-1mm}

\subsection{Discussions with the closely-related method}
We note that the recent method MAXIM~\cite{tu2022maxim} proposes a multi-axis MLP based architecture to solve imgae deraining. Different from MAXIM that employs the multi-axis gated strategy, our method utilizes a simple yet effective dimension rearrange mechanism to capture spatial information. First, we report the model complexity in Table \ref{table6}. Compared to MAXIM, our UDR-Mixer-L achieves a 81.6\% reduction in model parameters while decreasing FLOPs by 69.9\%. Note that as the training code of MAXIM is not available, we do not benchmark this approach on our proposed 4K-Rain13k. Since the testing code of MAXIM and the pre-trained model on the Rain13k dataset~\cite{jiang2020multi} are available, we compare their generalization ability of MAXIM and our method in real rainy scenes. As shown in Figure \ref{fig6} (b) and (c), our method successfully removes rain streaks and generates a clearer image. 

\section{Concluding Remarks}
This paper explores the task of UHD image deraining for the first time and proposes a high-quality dataset 4K-Rain13k to facilitate the performance comparison. Furthermore, we develop an efficient MLP-based method UDR-Mixer for UHD image deraining. Our approach utilizes a dimension rearrange mechanism to establish the global spatial context of UHD images and combines it with the original frequency representation of UHD images to help image restoration. The benchmark results show that our model achieves a favorable trade-off between performance and model complexity. 

\vspace{-2mm}

{\flushleft\textbf{Limitations}.} Although our method achieves favorable performance, it fails to handle the presence of fog-like rain accumulation in real rainy scenes. Future work will consider expanding 4K data with veiling effect and introducing physical models to guide enhancing the quality of image reconstruction.

{\small
\bibliographystyle{plainnat}
\bibliography{main}
}


\end{document}